\documentclass[journal]{IEEEtran}
\usepackage{cite}
\usepackage{amsmath}
\usepackage{multirow}
\usepackage{color}
\usepackage[switch]{lineno}
\usepackage{graphicx}
\usepackage{subfigure}
\usepackage{url}
\usepackage{epstopdf}

\begin{document}
%
\title{DISC: Deep Image Saliency Computing via Progressive Representation Learning}
%
%
%

\author{Tianshui~Chen,
        Liang~Lin,
        Lingbo~Liu,
        Xiaonan~Luo,
        and~Xuelong~Li
\thanks{This work was supported in part by the National Natural Science Foundation of China under Grant 61320106008 and 61232011, in part by Guangdong Natural Science Foundation under Grant S2013050014548, and in part by Guangdong Science and Technology Program under Grant 2013B010406005 and 2015B010128009. (Corresponding author: Liang Lin).
}
\thanks{T. Chen, L. Lin, L. Liu, and X. Luo are with the School of Data and Computer Science, Sun Yat-sen University, Guangzhou 510006, China. E-mail: linliang@ieee.org.}
\thanks{X. Li is with the Center for OPTical IMagery Analysis and Learning (OPTIMAL), State Key Laboratory of Transient Optics and Photonics, Xi'an Institute of Optics and Precision Mechanics, Chinese Academy of Sciences, Xi'an 710119, Shaanxi, P. R. China. E-mail: xuelong\_li@opt.ac.cn.}}

\maketitle

\begin{abstract}
Salient object detection increasingly receives attention as an important component or step in several pattern recognition and image processing tasks. Although a variety of powerful saliency models have been intensively proposed, they usually involve heavy feature (or model) engineering based on priors (or assumptions) about the properties of objects and backgrounds. Inspired by the effectiveness of recently developed feature learning, we provide a novel Deep Image Saliency Computing (DISC) framework for fine-grained image saliency computing. In particular, we model the image saliency from both the coarse- and fine-level observations, and utilize the deep convolutional neural network (CNN) to learn the saliency representation in a progressive manner. Specifically, our saliency model is built upon two stacked CNNs. The first CNN generates a coarse-level saliency map by taking the overall image as the input, roughly identifying saliency regions in the global context. Furthermore, we integrate superpixel-based local context information in the first CNN to refine the coarse-level saliency map. Guided by the coarse saliency map, the second CNN focuses on the local context to produce fine-grained and accurate saliency map while preserving object details. For a testing image, the two CNNs collaboratively conduct the saliency computing in one shot. Our DISC framework is capable of uniformly highlighting the objects-of-interest from complex background while preserving well object details. Extensive experiments on several standard benchmarks suggest that DISC outperforms other state-of-the-art methods and it also generalizes well across datasets without additional training. The executable version of DISC is available online: \url{http://vision.sysu.edu.cn/projects/DISC}.
\end{abstract}


\IEEEpeerreviewmaketitle

\section{Introduction}

\IEEEPARstart{A}{s} psychophysical experiments suggested, humans appear to perceive surrounding environment almost effortlessly due to their attentional mechanisms guiding the gaze to salient and informative locations in the visual field. Mimicking such a visual saliency system is a long-standing research topic both in neuroscience~\cite{itti2001computational} and in computer vision. Recently, instead of predicting sparse human eye fixation, many studies in computer vision focus on detecting the most informative and attention-grabbing regions (i.e., salient objects) in a scene~\cite{borji2012salient}. These proposed salient object detection methods~\cite{li2013saliency,yang2013saliency,wang2013saliency,wang2015pisa} also evolve to target on uniformly highlighting pixel-accurate saliency values, which is the aim of this paper. In addition, salient object detection has the great potential to benefit a wide range of applications, ranging from image/video compressing~\cite{christopoulos2000jpeg2000} and editing~\cite{han2006unsupervised} to object segmentation and recognition~\cite{lin2015discriminatively}.

\begin{figure*}[!t]
\centering
\includegraphics[width=0.9\linewidth]{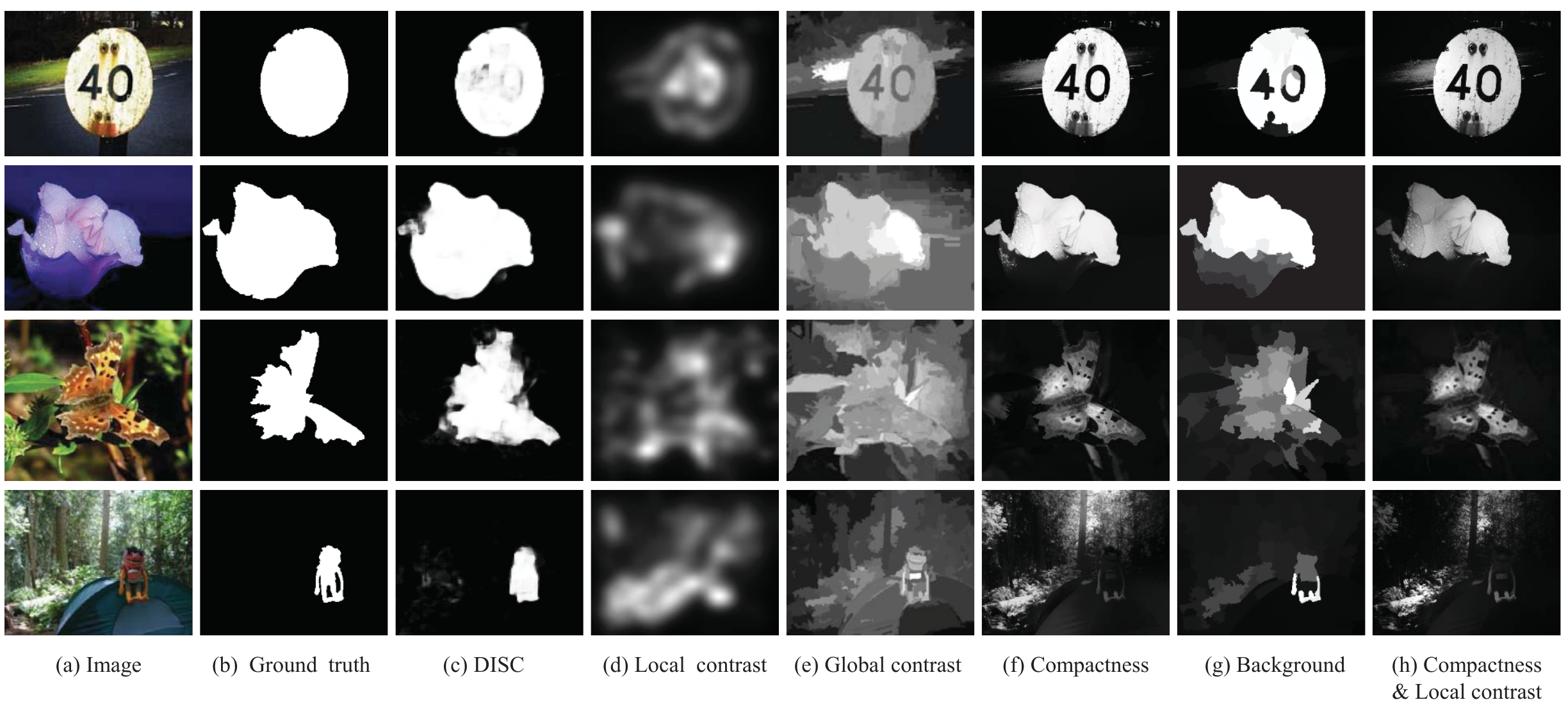}
\vspace{-10pt}
\caption{Comparison of image saliency maps generated by several state-of-the-art approaches based on different assumptions, including local/global contrast, compactness, background priors, and the combination of them. In contrast, the proposed method is capable to produce better fine-grained saliency maps without any assumption or feature engineering.}
\vspace{-20pt}
\label{fig:priors}
\end{figure*}

Due to the lack of a rigorous definition of image saliency, inferring the accurate saliency assignment for diversified natural images without a task orientation is a highly ill-posed problem. Therefore, many works of image saliency detection usually rely on various priors (or assumptions) for defining their saliency representations. Figure \ref{fig:priors} gives some examples. The contrast prior is arguably the most popular one, which can be further categorized as local contrast and global contrast according to the context where the contrast is computed. Local contrast based methods~\cite{itti1998model,liu2011learning} exploit pixel/patch difference in the vicinity to compute the image saliency. Without considering the global information, these methods, however, often miss the interior content while emphasizing the boundaries of salient objects, as shown in Figure \ref{fig:priors}(d). In contrast, global contrast based methods~\cite{zhai2006visual,achanta2009frequency} take the whole image as input to estimate the saliency of every pixel or image patch. Some results generated by these methods are shown in Figure \ref{fig:priors}(e). Though the entire salient objects are generally highlighted, the object structure details may not be well preserved. The compactness prior~\cite{perazzi2012saliency} is also widely utilized in image saliency modeling, which suggests the elements of salient objects tend to be compactly grouped in the image domain. This prior is shown to better capture object details compared to the global contrast, but may fail to highlight the object uniformly, as the examples presented in Figure \ref{fig:priors}(f). The background prior~\cite{yang2013saliency}, similar with the compactness, tends to render high saliency values to the regions near the center of the image. This prior, however, may lead to unreliable results on detecting salient regions that have similar appearance to the background, as shown in Figure \ref{fig:priors}(g). Some studies used a combination of different priors (e.g., the compactness and local contrast priors) to improve the performance, as shown in Figure \ref{fig:priors}(h).

Furthermore, the priors for modeling image saliency can be derived from higher-level knowledge. For example, the human perception suggests that red colored objects are more pronounced as more than 50\% of the cones in human eyes are red-sensitive~\cite{shen2012unified}. Some semantic information (e.g., object categories and attributes) has also been explored for completing the saliency models. The integration of these higher-level priors, nevertheless, is usually ad hoc, and one of the common ways is to calculate the weighted average of the saliency maps generated from high-level cues and those from low-level features.

On the other side, several learning-based approaches have been discussed for the task-driven image saliency detection. For instance, Liu et al.~\cite{liu2011learning} learned a conditional random field with supervision to detect salient objects. Jiang et al.~\cite{jiang2013salient} proposed to learn the saliency maps using a random forest regressor. Though impressive results are achieved, those methods often depend on hand-crafted features such as contrast histogram and color spatial distribution, and the design of these features basically follows the priors as we discussed above. Besides, most of the mentioned approaches conduct the saliency learning based on the over-segmentation of images (i.e., small regions or superpixels), and thus require an additional post-relaxation step (e.g., local filtering) to smooth the saliency values over pixels. The step of image segmentation may introduce errors and degenerate the saliency detection.

Different from those previous works, we present a novel end-to-end framework for fine-grained image saliency computing, and formulate it as a progressive representation learning problem. Our proposed model captures the image saliency in a coarse-to-fine manner, i.e., exploiting saliency cues with global and local contexts, respectively. Specifically, the coarse-level image saliency roughly identifies the locations, scales, and rough regions for the salient objects, while the object details (e.g., boundaries and subtle structures) will be rendered by the fine-level saliency map. Instead of defining two level representations with assumptions or hand-crafted image descriptors, we aim to learn the feature transformation directly from raw image pixels.

Inspired by its outstanding performance on traditional classification and detection tasks~\cite{krizhevsky2012imagenet}, we propose a novel image saliency model named DISC (i.e., Deep Image Saliency Computing) using convolutional neural network (CNN) in this paper. In particular, our DISC framework is built upon two stacked CNNs to cope with the coarse-to-fine saliency representation learning, where the CNN is treated as a feature extractor and the saliency map is generated by an additional linear transformation. The CNN architecture, designed according to the AlexNet~\cite{krizhevsky2012imagenet}, comprises several convolutional layers and fully connected layers. We define the linear transformation in the form of support vector machine (SVM) classification, instead of using a soft-max classifier like other CNN-based approaches. The combination of CNNs and SVMs has been discussed by Huang and LeCun~\cite{huang2006large}, but they trained these two models separately. In contrast, we embed the SVM into the CNN, ensuring joint optimization for these two components.

We briefly introduce the implementation of our approach as follows. The first CNN takes the whole image as input and measures the saliency score of each pixel in a global context, generating a coarse-level saliency map in a lower resolution. However, since the coarse CNN considers the whole image but pays less attention to local context information, it may mistakenly highlight some background regions or lose subtle salient structures. As a result, an incorrect coarse map will adversely affect the subsequent fine-level map generated by the second CNN. To address this issue, we utilize the superpixel-based local context information (SLCI) to further refine the coarse maps, which helps to keep the consistency of spatial structure of the salient objects. The SLCI comprises two components, namely intra-superpixel smoothing and inter-superpixel voting that make use of local context information in different scales. We formulate them as two special types of pooling layers and then embed them in the first CNN rather than treating them as a post-processing step. The second CNN is guided by the coarse-level saliency map, measuring the accurate and fine-grained saliency in a local context. Specifically, each pixel is fed to the second CNN together with its local observed patches from both the original image and the coarse-level saliency map. To avoid repeat computation, we load the entire image at a time, and thus make the neighboring pixels share their observations during the learning and inference procedures.
Moreover, we introduce a nonparametric map as an additional input channel to both CNNs, implicitly taking the spatial regularization into account to alleviate overfitting, and show that reasonable performance improvement can be achieved. Intuitively, we also refer the first and the second CNNs as the coarse-level and the fine-level CNNs, respectively.
The two CNNs are trained separately with supervision, yet they collaboratively conduct the inference for a testing image, i.e., producing the fine-grained saliency map in one shot. Some saliency maps generated by our approach are shown in Figure \ref{fig:priors}(c).

This paper makes three main contributions to the community.

\begin{itemize}
\item[\(\bullet\)] It presents a novel architecture capturing image saliency via progressive representation learning. This model is general to be extended to similar tasks such as scene parsing.
\item[\(\bullet\)] Superpixel-based local context information is integrated in the proposed framework for salient object structure preserving. It is formulated as two operations and embedded in the first CNN as intra-superpixel smoothing and inter-superpixel voting layers.
\item[\(\bullet\)] Extensive experiments on the standard benchmarks of image saliency detection demonstrate that our proposed method significantly outperforms state-of-the-art approaches and generalizes well across datasets without additional training. We also evaluate carefully each component of our model, and discuss the key components that improve the performance.
\end{itemize}

The remainder of the paper is organized as follows. Section 2 presents a review of related work. We then present our DISC model in Section 3, followed by a description of its learning algorithm in Section 4. The experimental results, comparisons and analysis are exhibited in Section 5. Section 6 concludes this paper.

\vspace{-6pt}
\section{Related Work}

According to the objective and the technical components of this work, we review the related work into three pipelines: image saliency detection, deep representation learning and deep saliency computing.

\vspace{-12pt}
\subsection{Image Saliency Detection}

Existing work of image saliency detection can be broadly divided into two categories: bottom-up and top-down approaches.

The bottom-up saliency models mainly focus on explaining visual attention according to different mathematical principles or priors. Among them, various contrast-based methods have been intensively discussed. Itti et al.~\cite{itti1998model} presented the center-surround operators based on the multi-scale image segmentation. Zhai and Shah \cite{zhai2006visual} proposed to use image histograms to compute pixel-level saliency map. Achanta et al. \cite{achanta2009frequency} provided a frequency tuned method that directly computed pixel saliency by subtracting the average image color. Cheng et al. \cite{cheng2011global} extended image histogram to 3D color space, and proposed color histogram contrast (HC) and region contrast (RC). Background information is also widely explored for saliency modeling, including the boundary prior and the background-connectivity prior~\cite{yang2013saliency}. The compactness prior encourages the salient elements to be grouped tightly, and it was realized by Perazzi et al.~\cite{perazzi2012saliency} using two Gaussian filters. Moreover, several mathematical models have been also utilized to define the bottom-up saliency models, such as entropy \cite{kadir2001saliency} and the Shannon's self information \cite{bruce2005saliency}. Shen and Wu~\cite{shen2012unified} proposed to utilize the low-rank representation for saliency detection, which is based on the assumption that the non-salient background usually lies in a low-dimensional subspace while the sparse noise indicates the salient regions.

The top-down approaches introduce visual knowledge commonly acquired through supervised learning to detect image saliency. Approaches in this category are highly effective on task-specified saliency detection. For example, Lin et al. \cite{lin2014computational} proposed a computational visual saliency model based on feature-prior, position-prior, and feature-distribution, which are learned from support vector regressor (SVR), ground truth of training images and features in the image using information theory, respectively. Mai et al. \cite{mai2013saliency} trained a conditional random field (CRF) model to aggregate various saliency map produced by different methods. Lu et al. \cite{lu2014learning} proposed a graph-based method to learn optimal seeds for object saliency. They learned the combination of different features that best discriminate between object and background saliency.

\begin{figure*}[tp]
\centering
\includegraphics[width=0.9\linewidth]{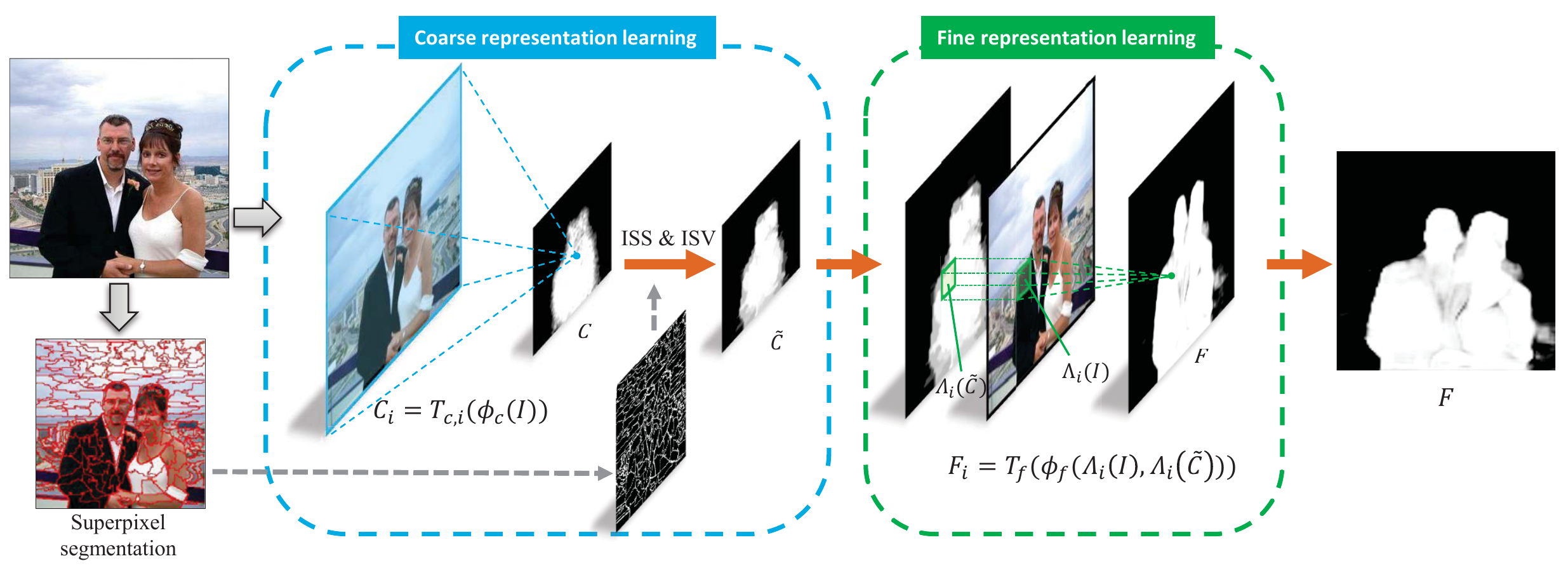}
\vspace*{-6pt}
\caption{Illustration of our proposed deep saliency computing model. The first CNN takes the whole image data as input and produces coarse map. Guided by the coarse map, the second CNN takes a local patch as input and generates the fine-grained saliency map.}
\vspace*{-20pt}
\label{fig:architecture}
\end{figure*}

\vspace{-10pt}
\subsection{Deep Representation Learning}
Recently, representation learning via deep CNNs has obtained great success in various of computer vision tasks, such as image classification \cite{krizhevsky2012imagenet}, object detection \cite{liang2015towards}, Person Re-identification \cite{ding2015deep,deephashing} and human centric analysis \cite{liang2015human,lin2015deepmodel}.
Many works also apply multi-scale deep network to various computer vision tasks. Farabet et al. \cite{farabet2013learning} used a multi-scale CNN trained from raw pixels to extract dense feature for assigning a label to each pixel. However, it is time-consuming for multiple complex post-processing required for accurate prediction. These works utilized multi-scale input to realize multi-scale representation learning. Instead, some works integrated multi-scale structure inside the CNN. Sermanet et al. \cite{sermanet2013pedestrian} incorporated multi-scale information inside the CNN architecture and utilized unsupervised multi-stage feature learning for pedestrian detection. It produced features that extract both global structures and local details, and improved the performance by a large margin. However, this novel architecture was difficult to adapt to high-resolution pixel-wise labeling. Nowadays, more works tended to design different network architecture for different scale representation learning. Wang et al. \cite{wang2014deep} designed a localization network and a segmentation network to rapidly locate and accurately segment the object from the images. It outperformed previous segmentation work on public benchmark both in accuracy and efficiency. However, it heavily relied on the accuracy of localization, which is not always satisfied, especially for the image with more than one object. Sun et al. \cite{sun2013deep} designed a cascaded regression framework with three level of CNNs for facial landmark detection. Eigen et al. \cite{eigen2014depth} proposed a depth prediction model based on two deep CNNs to estimate the depth of each image pixel in a coarse-to-fine manner. The output of the first CNN was concatenated with the output of the first convolutional layer of the second CNN and they are fed into the subsequent layers for accurate depth prediction.

\vspace{-10pt}
\subsection{Deep Learning for Saliency Detection}

Very recently, deep neural network models are also utilized in image saliency detection \cite{he2015supercnn, li2015visual, wang2015deep}. For example, He et al. \cite{he2015supercnn} learned the hierarchical contrast features using CNNs and designed a multi-scale architecture with sharing weights for robust salient region detection. Li et al. \cite{li2015visual} extracted multi-scale deep features using CNNs pre-trained on the ImageNet dataset \cite{russakovsky2014imagenet}, and the multi-scale representations were fused to produce the final saliency score by several fully connected layers. Wang et al. \cite{wang2015deep} presented a saliency detection algorithm by integrating both local estimation and global search, and utilized two deep neural networks to realize it. Some other attempts have been made for the application of human fixation prediction \cite{liu2015predicting, vig2014large, kummerer2014deep, shen2014learning}. Vig et al. \cite{vig2014large} combined CNNs and a linear SVM for predicting fixation, and Deep Gaze I  \cite{kummerer2014deep} learned a weight on linear combination of convolution channels without using fully connected layers. Using deep architectures, these above mentioned works mainly integrated multi-scale context information in a straightforward way (e.g., in parallel). In contrast, we develop our saliency model following a divide-and-conquer manner, i.e., progressively generating saliency maps, which finely accords with biological perception and existing coarse-to-fine object detection models. We also demonstrate the superior performance of our method over other existing deep saliency model.

\vspace{-6pt}
\section{Deep Image Saliency Computing Framework}

In this section, we describe the proposed DISC framework in detail. We model the fine-level image saliency computing as a progressive representation learning problem with two stacked CNNs. The first CNN takes the original image as input and produces a coarse-level saliency map. Both the original image and the coarse-level saliency map are then fed into the second CNN, generating the final fine-level saliency map. To maintain the spatial structure of salient object for the coarse map, the superpixel-based local context information is integrated in the first CNN as an intra-smoothing layer and an inter-voting layer. The pipeline of DISC is illustrated in Figure \ref{fig:architecture}.

\begin{figure*}[!t]
\centering
\includegraphics[width=0.95\linewidth]{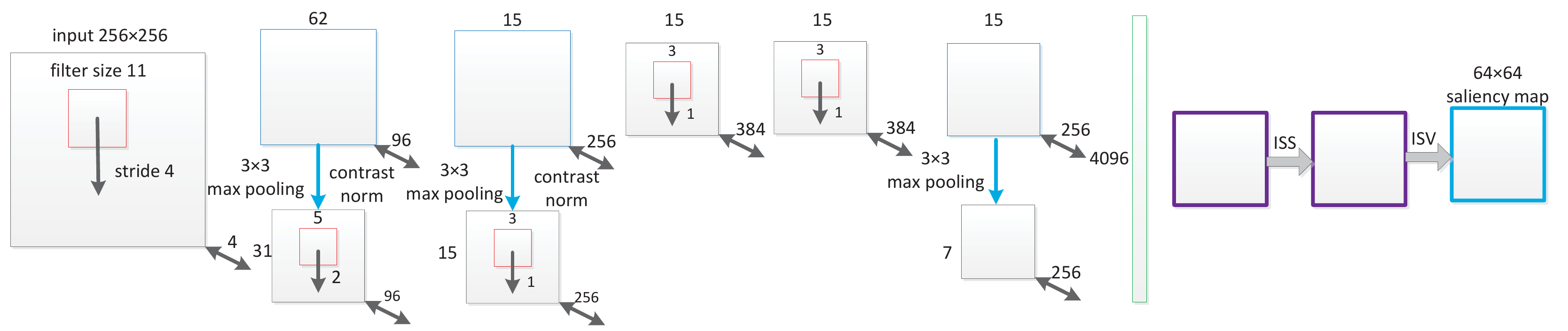}
\vspace{-6pt}
\caption{Architecture of the first CNN for coarse-level saliency computing. A $256\times256$ image is convolved with 96 different 1st layer filters, with kernel size of $11\times11$ and a stride of 4. The obtained feature maps are then passed through a rectified linear function, max pooling operation, and local contrast normalizations and similar operations are repeated in the next four convolutional layers. The feature vector ${\phi _c}(I)$ of the whole image is then fed into the fully connected layer that produces the $64\times64$ coarse-level saliency map. This map is further refined by ISS and ISV.}
\vspace{-20pt}
\label{fig:globalnet}
\end{figure*}

\vspace{-12pt}
\subsection{Progressive Representation Learning}

Before delving into the formulation, we present some notations that would be used later. Let $I$, $C$, $F$ denote the original image data, coarse-level and fine-level saliency maps, respectively; $\Lambda_i(I)$ and $\Lambda_i(C)$ denote the square patches of $I$ and $C$ centered at pixel $i$; $C_i$ and $F_i$ denote the saliency score of pixel $i$ in saliency maps $C$ and $F$.

The coarse-level saliency map is generated first. We extract the feature of the whole image, and then apply a linear transformation that assigns a corresponding saliency score to each pixel, which can be expressed as

\begin{equation}
\begin{aligned}
{C_i}&={\mathcal{T}_{c,i}}\left({\phi_c}(I)\right)\\
&=\mathbf{w}_{c,i}^T{\phi_c}(I) + {b_{c,i}},\ i=1,2,\dots,N_c,
\label{eqn:coarse_rep}
\end{aligned}
\end{equation}
where $\mathbf{w}_{c,i}$ and $b_{c,i}$ are the parameters of the linear transform $\mathcal{T}_{c,i}$ for pixel $i$. $N_c$ is the pixel number of the coarse-level saliency map. The feature extractor $\phi_c$ is implemented by a CNN, and the linear transformations $\{\mathcal{T}_{c,i}\}_{i=1}^{N_{c}}$ are defined in the form of linear SVMs. We further refine $C$ using superpixel-based local context information (SLCI) to get a structure-preserved coarse map $\tilde{C}$. We describe SLCI in detail in the next subsection.

To compute the saliency score of each pixel $i$ in the fine saliency map, we take both the local patch $\Lambda_i(I)$ from the original image and its corresponding patch $\Lambda_i(\tilde{C})$ from the coarse-level saliency map as input, and map them to a feature vector ${\phi_f}\left(\Lambda_i(I),\Lambda_i(\tilde{C})\right)$. A simple linear transformation $\mathcal{T}_f$ is then utilized to map the feature vector to the corresponding saliency score.

\vspace{-6pt}
\begin{equation}
\begin{aligned}
{F_i}&={\mathcal{T}_f}({\phi_f}(\Lambda_i(I),\Lambda_i(\tilde{C}))) \\
&=\mathbf{w}_f^T{\phi_f}(\Lambda_i(I),\Lambda_i(\tilde{C})) + {b_f},\ i=1,2,\dots,N_f,
\label{eqn:fine_rep}
\end{aligned}
\end{equation}
where $\mathbf{w}_f$ and ${b_f}$ are the parameters of the linear transformation ${\mathcal{T}_f}$. $N_f$ is the pixel number of the fine-level saliency map. Similarly, the feature extractor $\phi_f$ is implemented by another CNN. For each CNN, We jointly train the feature extractor and the classifier for the saliency computing model.

\vspace{-12pt}
\subsection{Superpixel-based Local Context Information (SLCI)}
\label{subsec:SLCI}

As discussed above, the fine-level CNN generates the final saliency maps guided by the coarse maps. Recall that the saliency score of each pixel in the fine map is influenced merely by its small neighbourhood from the original image and the corresponding coarse map. Hence, the accuracy of fine saliency map depends heavily on the quality of coarse saliency map. Note that the first CNN takes the global information into consideration, but pays less attention to the nearby local context. We experimentally find that although the coarse CNN is able to highlight the overall salient regions, it suffers from two main problems:~(1) the generated coarse maps may confuse some of small foreground or background regions if they have similar appearance;~(2) it may fail to preserve the structure of the salient objects especially when the background is very cluttered. Our aim is to capture the global information and simultaneously consider nearby context in order to produce high-quality coarse maps for the subsequent learning. Therefore, the SLCI, consisting of two types of superpixel-level refinements called intra smoothing and extra voting, is utilized to make better use of local context information and preserve the spatial structure information. They are integrated in the first CNN as two pooling layers for coarse saliency map prediction. Note that the fine CNN predicts the saliency score of each pixel according to its neighborhood, and local context has been fully considered here, so we do not need to use SLCI for the fine CNN.

\noindent \textbf{Intra-superpixel smoothing (ISS).}
ISS aims to assign close saliency scores to the pixels with similar appearance in local small regions.
We first over segment the input image using the entropy rate based segmentation algorithm \cite{liu2011entropy} to obtain $N$ small superpixels per image. Let $R_i$ denote the superpixel that contains pixel $i$. Given the prediction map $C$, the intra-smoothed saliency map $\tilde C$ can be computed by

\begin{equation}
{\tilde C_i} = \frac{1}{| {{R_i}} |}\sum\limits_{j \in {R_i}} {{C_j}},
\label{eqn:intra_smooth}
\end{equation}
where ${| \cdot |}$ is the cardinality of the set. In this way, the saliency scores of pixels within a superpixel are replaced by the average score.

\noindent \textbf{Inter-superpixel voting (ISV).}
ISS focuses on saliency score smoothing in small regions, which can deal with small-scale saliency structure. Further, ISV takes larger regions into account for preserving large-scale saliency structure, so that the salient object can be labeled more uniformly. We replace the saliency score of each region by the weighted average of the saliency scores of the adjacent superpixels. Given a superpixel $s$, we first compute the LAB color histogram $h^{c}(s)$ and the gradient histogram $h^{g}(s)$, and then concatenate them to build the appearance feature histogram $h(s)$. Let $F_s$ denote the saliency value of $s$. The inter-voting can be expressed by
\begin{equation}
{\tilde C_s} = (1 - \lambda ){C_s} + \lambda\sum\limits_{s'\in D(s)}w(s')\cdot C_{s'},
\label{eqn:extra_voting}
\end{equation}
where the weight for $s'$ is defined as
\begin{equation}
w(s')=\frac{\exp{\left( { - \left\| {h(s) - h(s')} \right\|} \right)}}{{\sum\limits_{s'' \in D(s)}\exp{\left( { - \left\| {h(s) - h(s'')} \right\|} \right)}}},
\end{equation}
${D(s)}$ contains all superpixels that are adjacent to $s$, and $\|{h(s) - h(s')}\|$ is the Euclidean distance between feature histogram $h(s)$ and $h(s')$. $\lambda$ is a scaling factor to balance the two terms. The voting weights of each region are defined according to the similarity in the color and structure space, which are two important cues for saliency computing \cite{wang2015pisa}. We assign a larger voting weight to the region with similar color and structure feature to $s$.

It can be observed that ISS and ISV can be regarded as two special types of pooling methods. For ISS, the pooling map is calculated by averaging the predicted map in small regions with arbitrary shape. For ISV, we calculate the weighted average of the intra-smoothing response map in a larger region. We formulate these two operations as ISS-pooling and ISV-pooling layers, and then integrate them as sub-components in the first CNN. During the training stage, the gradients are computed according to Equation \ref{eqn:intra_smooth}, \ref{eqn:extra_voting}.

\begin{figure*}[!t]
\centering
\includegraphics[width=0.95\linewidth]{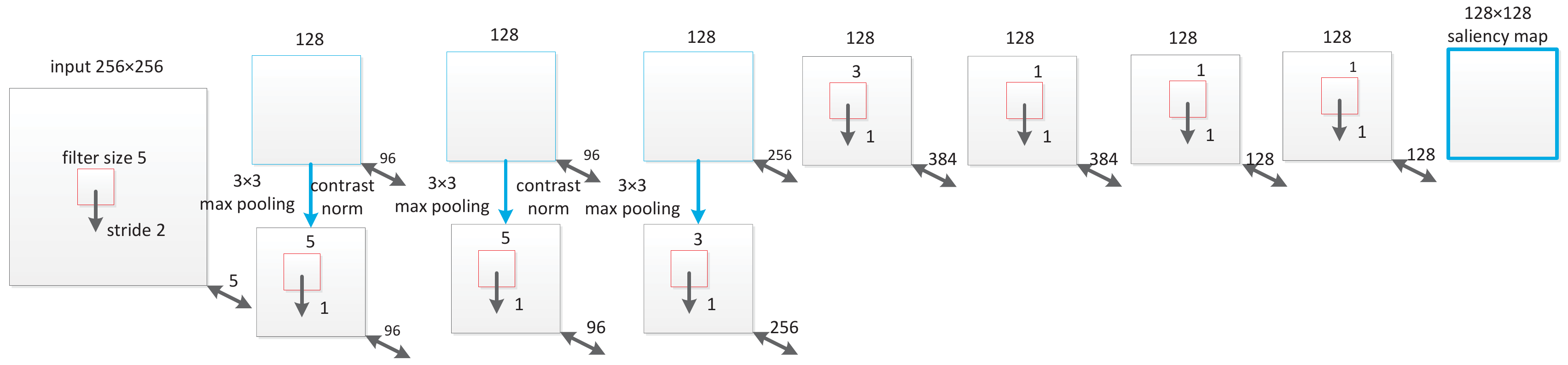}
\vspace{-6pt}
\caption{Architecture of the second CNN for fine-level saliency computing. The network takes $256\times256\times5$ image data as input and convolves the data with 96 different first layer filters, with kernel size of $5\times5$ and a stride of 2. The obtained feature maps are then passed through a rectified linear function, max pooling operation, contrast normalization. The next six convolutional layers take similar operations but with all stride of 1 and output the dense feature vectors ${\phi_f}(\Lambda_i(I),\Lambda_i(C)) (i=1,2,\dots,N_f)$. The dense feature vectors are fed into the the last convolutional layer, producing the $128\times128$ fine-grained saliency map.}
\vspace{-20pt}
\label{fig:localnet}
\end{figure*}

\vspace{-12pt}
\subsection{Spatial Regularization}
In this subsection, we introduce the spatial regularization, which helps to reject false label assignments and leads to an improvement of the performance. As suggested in \cite{liu2011learning}, objects near the center of the image are more likely to be salient. A mass of saliency detection algorithms such as \cite{wang2015pisa, shen2012unified, lu2014learning} incorporate it in their saliency computing frameworks as a center-bias prior. In contrast, we simply average all of the aligned ground truth saliency map in the training set and take it as an additional input channel in both training and testing procedures for spatial regularization. We do not use a center-bias prior here for two reasons. First, previous works formulate the center-bias prior as a parametric spatial term or a post-processing step, and inevitably have to tune the parameters carefully for better performance. Second, incorporating the center bias prior may suppress some salient regions if they appear near the image boundaries. Conversely, the spatial regularization map is considered as an additional input channel, and fed to the CNNs with three-channels original image data. In this way, the network is capable to learn saliency representation from the original images, while implicitly taking the spatial regularization into account to alleviate overfitting.

\vspace{-12pt}
\subsection{Network Architecture}
As discussed above, two stacked CNNs are adopted for the coarse-to-fine saliency representation learning. This subsection introduces the architectures of DISC.

The first CNN takes the whole image as input and produces the coarse-level saliency map. It contains five convolutional layers and one fully connected layer, followed by the ISS and ISV pooling layers, as illustrated in Figure \ref{fig:globalnet}. It is similar to the general architecture proposed by Krizhevsky et al. \cite{krizhevsky2012imagenet}, but we use four-channel data as input and replace the last two fully connected layers by ISS and ISV pooling layers. The input image data contain three RGB channels of the training image and one channel of the spatial regularization map. The five convolutional layers are served as the feature extractor, which takes the whole image data of size $256\times256\times4$ as input and produces a feature vector of length $7\times7\times256$. The last fully connected layer computes the linear transformations of the feature vector and outputs 4,096 saliency scores that are re-arranged to a coarse-level map in a lower resolution. The ISS-pooling and ISV-pooling layers sequentially refine the coarse maps.

The second CNN, guided by the coarse-level map, computes the saliency score for each pixel based on local observation. It is designed as a fully convolutional network \cite{long2015fully}, as shown in Figure \ref{fig:localnet}. The input data contains five channels, i.e., three channels of original images with two channels of the spatial regularization and coarse maps. This network contains eight convolutional layers and first seven of which are regarded as a feature extractor. Except for the first convolutional layer, we set the stride as 1 based upon the following considerations. First, the feature extractor can produce dense feature vectors for accurate saliency computing. Second, the local patch for saliency computing is kept small, ensuring to better preserve the local details. The last convolutional layer takes the feature vector of each pixel as input, and then computes its corresponding saliency value to form the fine-level saliency representation.

As the second CNN focuses on local context, it is intuitive to crop a local patch for each pixel and compute their saliency scores separately. However, it is time-consuming for network training and inference. For example, to get a $128\times128$ saliency map, 16,384 patches have to be computed. Note that most area of two adjacent patches is overlapping. To avoid these redundant computation, the CNN takes the whole image as input and produces dense outputs. In this way, the patches with overlap share much intermediate computation results, significantly reducing the training and inference time by hundreds of times.

\vspace{-6pt}
\section{Optimization}

The two CNNs are trained using the stochastic gradient descent (SGD) algorithm with hinge loss sequentially.

\vspace{-12pt}
\subsection{Optimization Formulation}
Suppose that there are $N$ training images, the training sets for the two CNNs are $\mathcal{X}_c=\{(I^k, Y^k)\}_{k=1}^N$ and $\mathcal{X}_f=\{(I^k, C^k, Y^k)\}_{k=1}^N$ respectively. $I^k$ is the four-channel image data, including three channels of RGB value and a channel of  center-bias map. $Y^k$ is the corresponding saliency map of the $k$-th image. $C^k$ is the coarse map produced by the first CNN, which would be fed into the second CNN. The optimization objective is to minimize the total loss

\vspace{-6pt}
\begin{equation}
\mathcal{L} = {\mathcal{L}_c} + {\mathcal{L}_f},
\label{eqn:loss_total}
\end{equation}
where ${\mathcal{L}_c}$ and ${\mathcal{L}_f}$ are the objective functions for the first and second CNNs respectively. We optimize the two terms separately.

For the first CNN, we formulate the coarse saliency computing for each pixel as a binary classification problem and utilize hinge loss to optimize the classifier. So ${\mathcal{L}_c}$ can be defined as

\vspace{-6pt}
\begin{equation}
\begin{aligned}
\mathcal{L}_c&=\frac{1}{2}\sum\limits_{i = 1}^{{N_c}} {\;\mathbf{w}_{c,i}^T{\mathbf{w}_{c,i}}} + \\
 &C\sum\limits_{i = 1}^{{N_c}} {\sum\limits_{k = 1}^N {\max (1 - Y_i^k(\mathbf{w}_{c,i}^T\phi_c ({I^k}) + {b_{c,i}}),0)^2} }.
\label{eqn:loss_cnn1}
\end{aligned}
\end{equation}
We adopt the hinge loss \cite{xu2013soft} in the square form for guaranteeing it is differential. According to Equation (\ref{eqn:loss_cnn1}), the gradients can be computed by

\vspace{-6pt}
\begin{equation}
\begin{aligned}
&\frac{{\partial {\mathcal{L}_c}}}{{\partial {\phi _c}({I^k})}} = \\
&- 2C\sum\limits_{i = 1}^{{N_c}} {\sum\limits_{k = 1}^N {Y_i^k{\mathbf{w}_{c,i}^T}\max (1 - Y_i^k(\mathbf{w}_{c,i}^T{\phi _c}({I^k}) + {b_{c,i}}),0)} }.
\label{eqn:gradient_cnn1}
\end{aligned}
\end{equation}
Similarly, the fine-level saliency computing can also be regarded as binary classification problem, and the objective function ${\mathcal{L}_f}$ is defined as

\begin{equation}
\begin{aligned}
\mathcal{L}_f&=\frac{1}{2}\mathbf{w}_f^T{\mathbf{w}_f}+ \\
 &C\sum\limits_{k = 1}^N {\sum\limits_{i = 1}^{{N_f}} {\max {{(1 - Y_i^k(\mathbf{w}_f^Tf_i + {b_f}), 0)}^2}} },
\label{eqn:loss_cnn2}
\end{aligned}
\end{equation}
where we denote ${\phi_f}(\Lambda_i(I),\Lambda_i(C))$ as $f_i$ for simplicity.
The corresponding gradients can be calculate by
\begin{equation}
\begin{aligned}
&\frac{{\partial {\mathcal{L}_f}}}{{\partial f_i}} = \\
&- 2C\sum\limits_{k = 1}^N {\sum\limits_{i = 1}^{{N_f}} {Y_i^k\mathbf{w}_f^T\max (1 - Y_i^k(\mathbf{w}_f^Tf_i + {b_f}),0)} }.
\label{eqn:gradient_cnn2}
\end{aligned}
\end{equation}

The parameters of lower layers are learned by backpropagating the gradients from the top layer, which can be computed by differentiating the objective function with respect to the activations of the penultimate layer, that is $\phi_c({I^k})$. According to Equations (\ref{eqn:loss_cnn1}, \ref{eqn:loss_cnn2}), the gradients can be expressed as (\ref{eqn:gradient_cnn1}, \ref{eqn:gradient_cnn2}). From this point on, the back propagation algorithm is exactly the same as the standard softmax-based deep networks.

We first remove the ISS and ISV layers from the first CNN and initial the parameters of the first CNN from a zero-mean Gaussian distribution. We then train it using stochastic gradient descent with backpropagation method. Then we integrate ISS and ISV layers in the first CNN and fine tune it to get the final coarse model. The parameters of the second CNN are initialized and trained identically.

\vspace{-6pt}
\section{Task-Oriented Adaptation}
\begin{figure}[htp]
\centering
\includegraphics[width=0.9\linewidth]{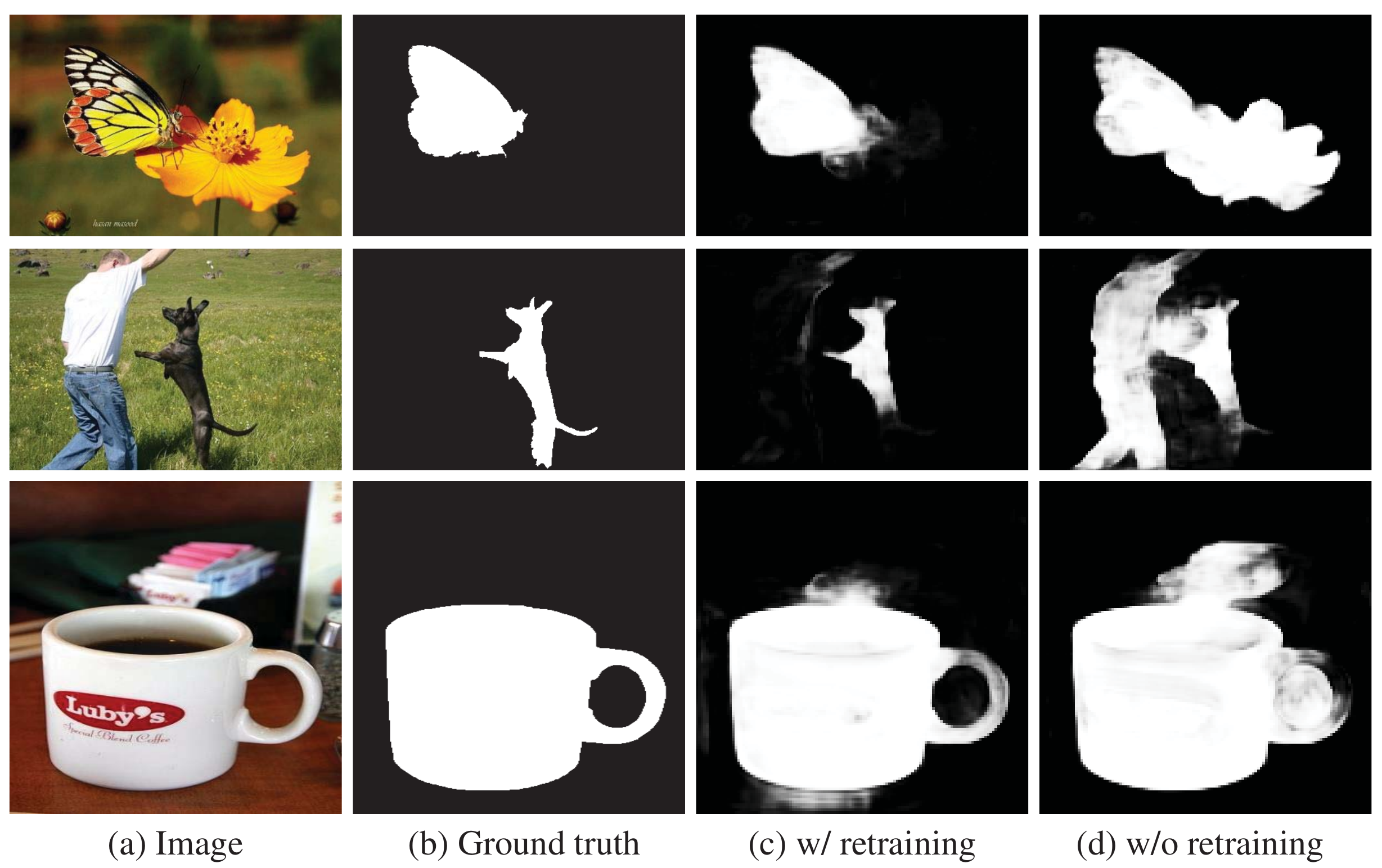}
\vspace{-6pt}
\caption{The result of task-oriented salient object detection without and with retraining. All the salient objects are highlighted in (d), but only the specific object is highlighted after retraining in (c).}
\vspace{-20pt}
\label{fig:task-oriented}
\end{figure}

Task oriented salient object detection targets on highlighting specific classes of salient objects. It is significant in the situation that we are interested merely in some classes of objects. As stated above, the model trained in a generic salient object dataset is inclined to highlight all the salient objects in the image, no matter what it is. Taking the first image in Figure \ref{fig:task-oriented} as an example, the model trained on the MSRA dataset highlights both the butterfly and the flower because both are salient. Hence, such a model cannot be directly used in a task-oriented task (e.g., only highlighting the butterfly).

Fortunately, it is easy to generalize the model to task-oriented task. We first collect a dataset of images that contain the specific classes of object that we are interested in. We then label the pixels that belong to these specific classes of object as 1 and others as -1 for each image. Finally, we fine tune the model on the collected dataset. In this way, only the specific kinds of objects are highlighted, as shown in Figure \ref{fig:task-oriented}.

\vspace{-6pt}
\section{Experiments}

\subsection{Experiment Setting}

\noindent \textbf{Dateset description.}
We evaluate our DISC framework on five public benchmark datasets: MSRA10K \cite{cheng2011global}, SED1 \cite{alpert2007image}, ECSSD \cite{yan2013hierarchical}, PASCAL1500 \cite{zou2013segmentation} and THUR15K \cite{cheng2014salientshape}. The MSRA10K contains 10,000 images from MSRA dataset with pixel-level labeling for salient objects. Because most images contain only a single object located near the center of the image and background is generally clean, the accuracy of recent methods has been more than 90\%, but our DISC model still improve the accuracy greatly. The SED1 dataset is exploited recently which contains 100 images of single objects. The ECSSD contains 1,000 diversified patterns in both foreground and background, which includes many semantically meaningful but structurally complex images for evaluation. The PASCAL1500, created from PASCAL VOC 2012 segmentation challenge, is also a challenging dataset, in which images contain multiple objects appearing at a variety of locations and scales with complicated background. THUR15K consists of 15,000 images, with 6,233 pixel-accurate ground truth annotations for five specific categories: butterfly, coffee mug, dog jump, giraffe, and plane.

\begin{figure*}[!t]
\centering
{\includegraphics[width=0.8\linewidth]{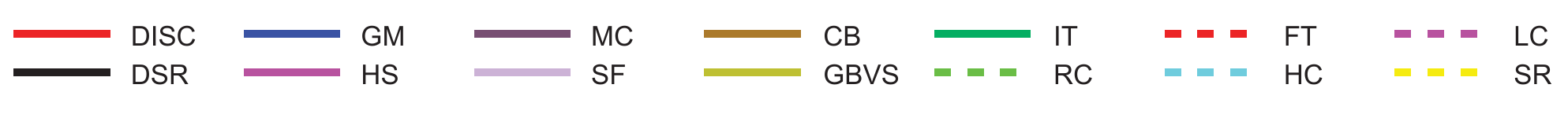}}
{\includegraphics[width=0.12\linewidth]{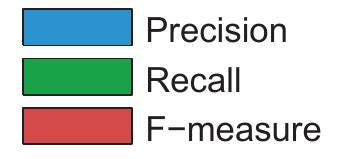}}
\subfigure[]{
\label{fig:subfig1_1} 
\includegraphics[width=0.32\linewidth]{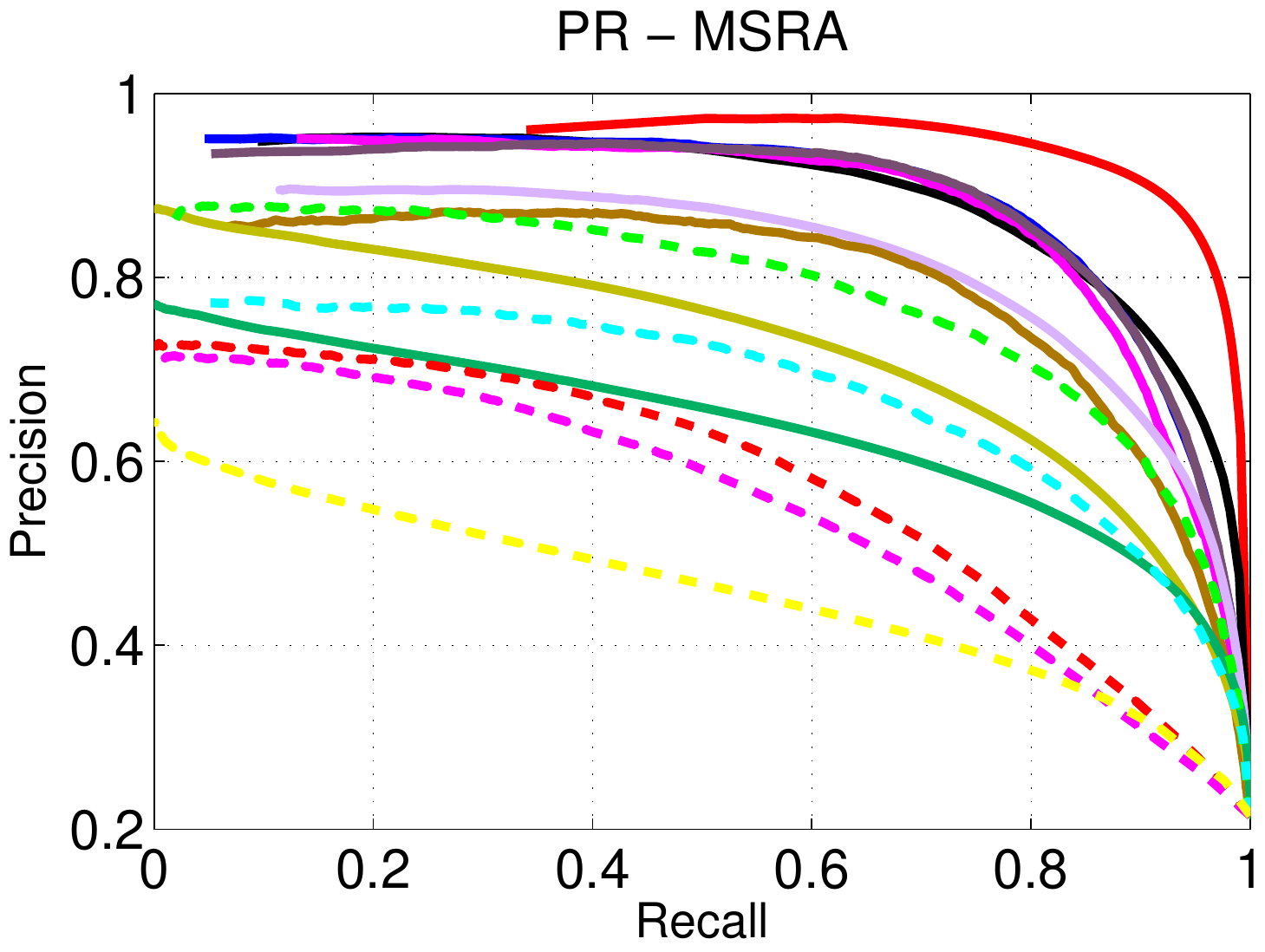}}
\subfigure[]{
\label{fig:subfig1_2} 
\includegraphics[width=0.32\linewidth]{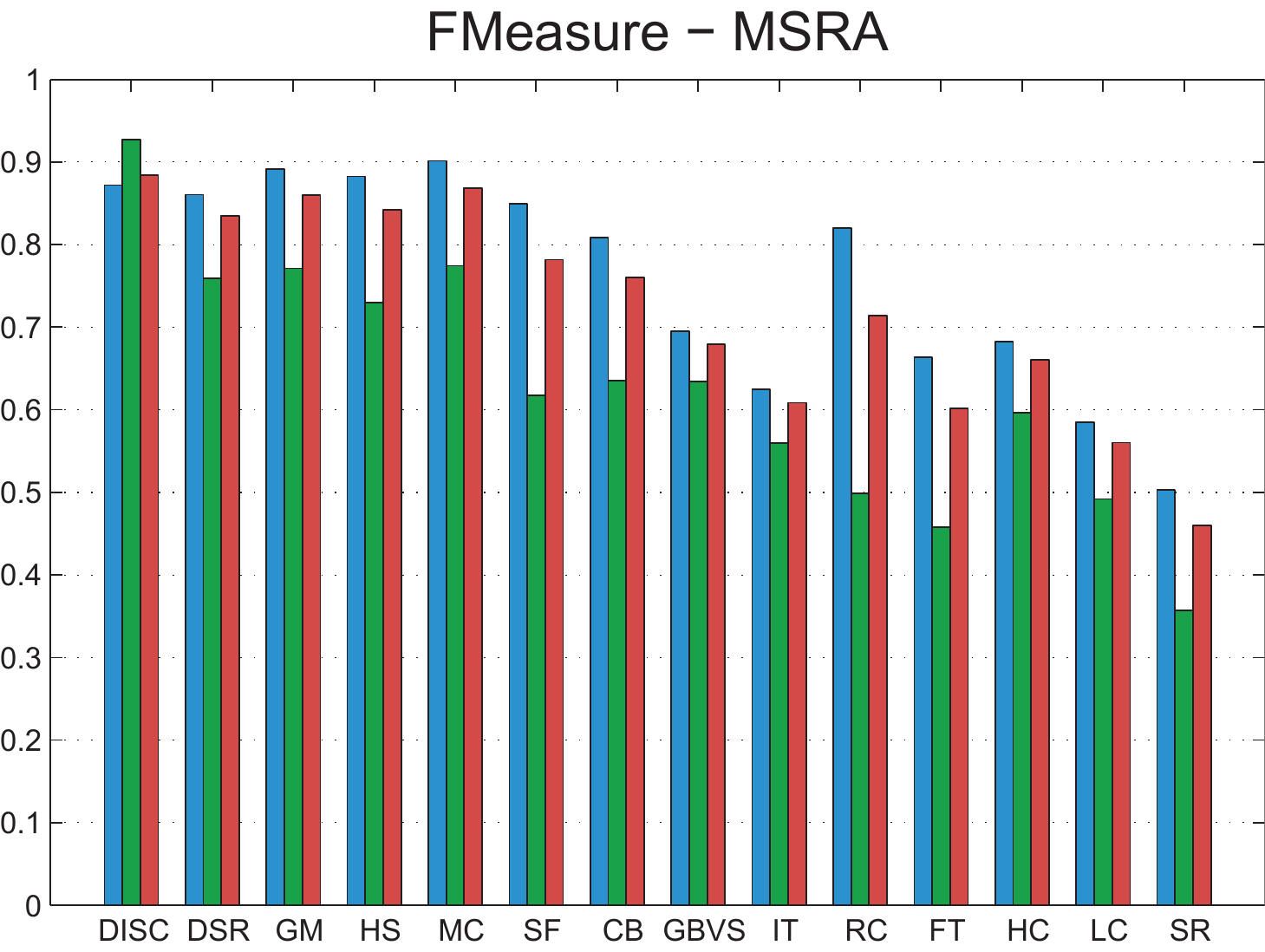}}
\subfigure[]{
\label{fig:subfig1_3} 
\includegraphics[width=0.32\linewidth]{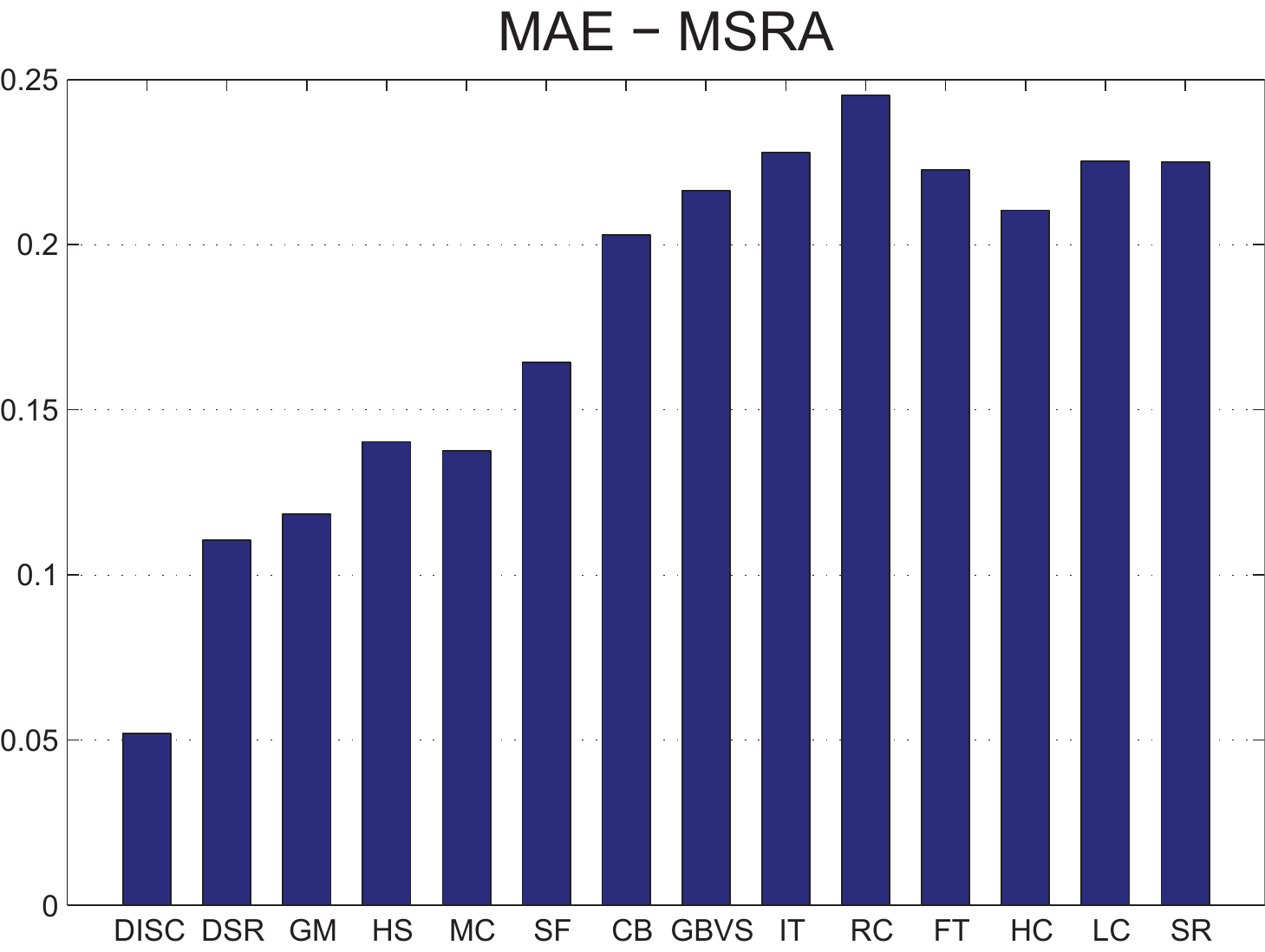}}
\caption{Experimental results on the MSRA10K dataset. (a) Precision-recall curve, (b) precision-recall bar with F-measure, and (c) mean absolute error for comparing our DISC model against previous works. Best viewed in color.}
\label{fig:result_MSRA}
\end{figure*}

\begin{figure*}[!t]
\centering
{\includegraphics[width=0.8\linewidth]{PR_Legend.pdf}}
{\includegraphics[width=0.12\linewidth]{FMeasure_Legend.pdf}}
{\includegraphics[width=0.325\linewidth]{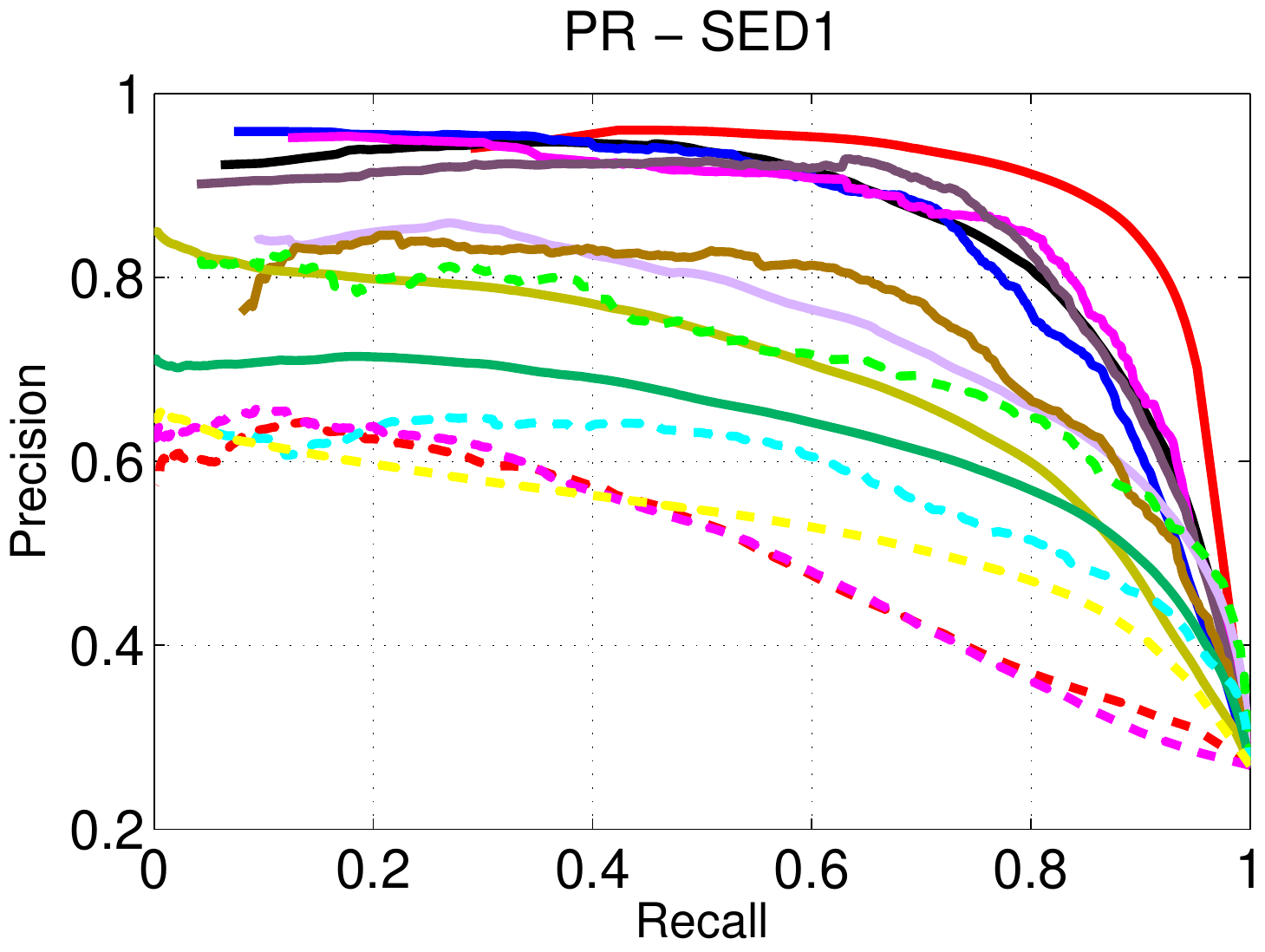}}
{\includegraphics[width=0.325\linewidth]{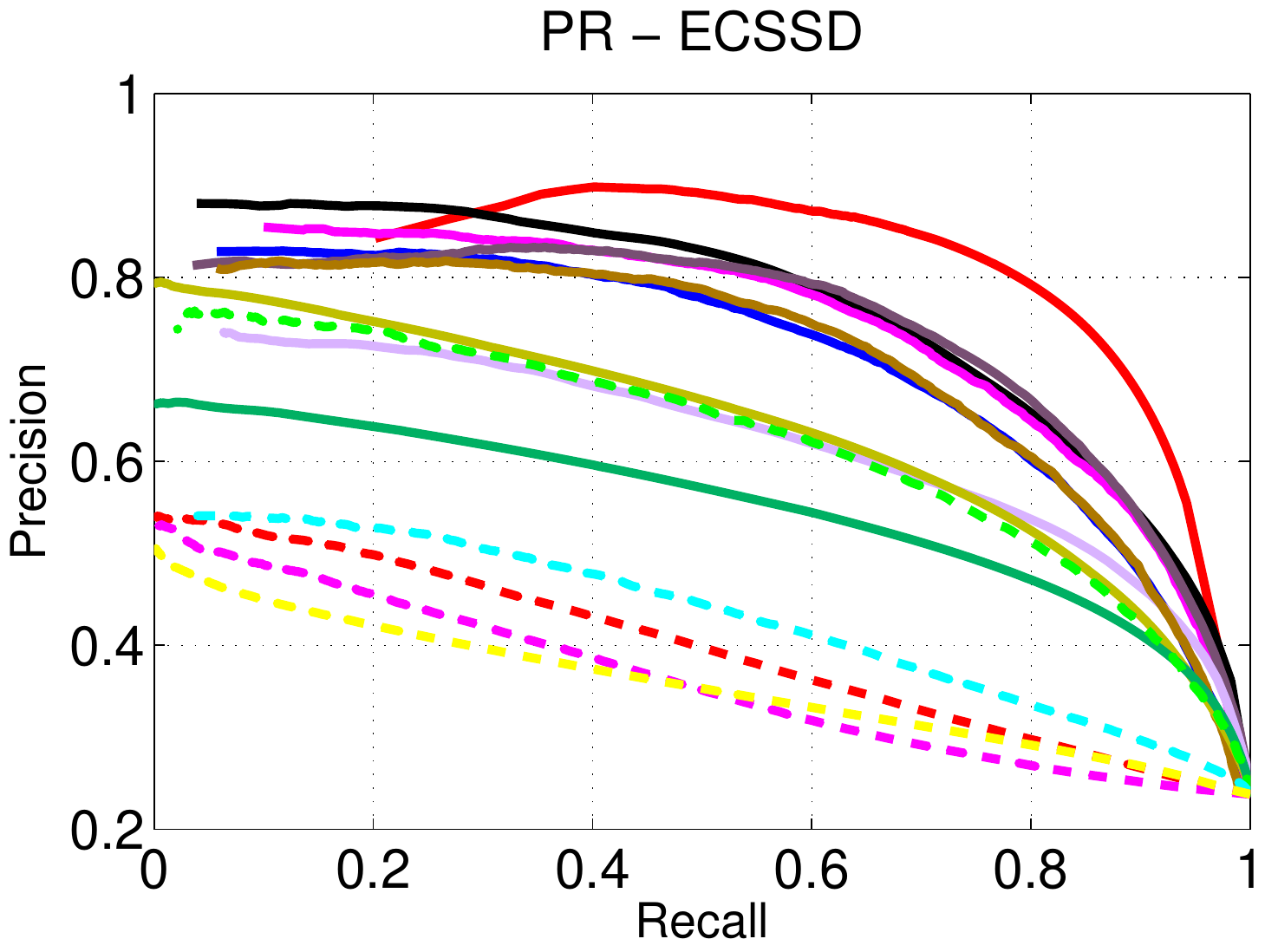}}
{\includegraphics[width=0.325\linewidth]{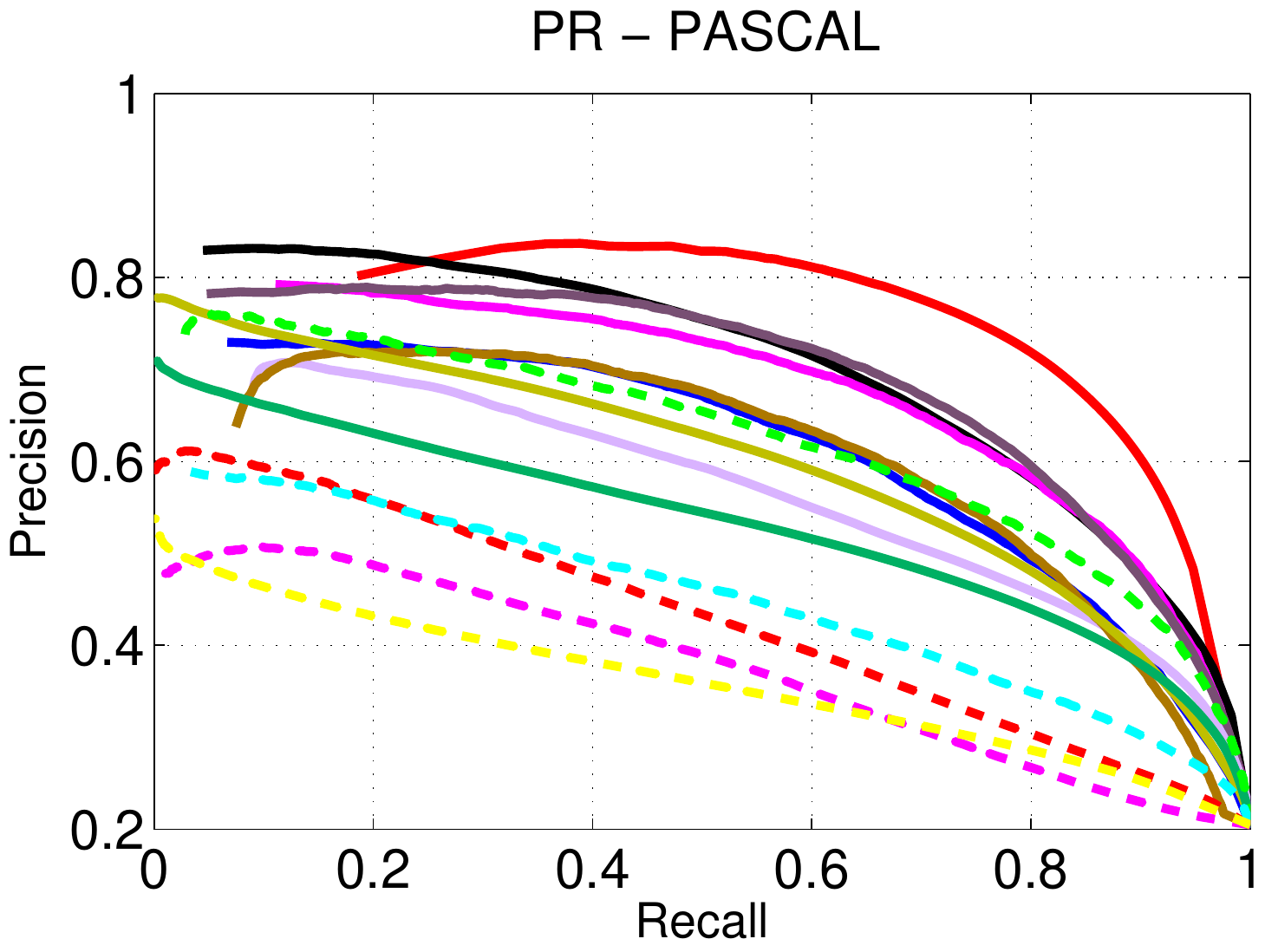}}
{\includegraphics[width=0.32\linewidth]{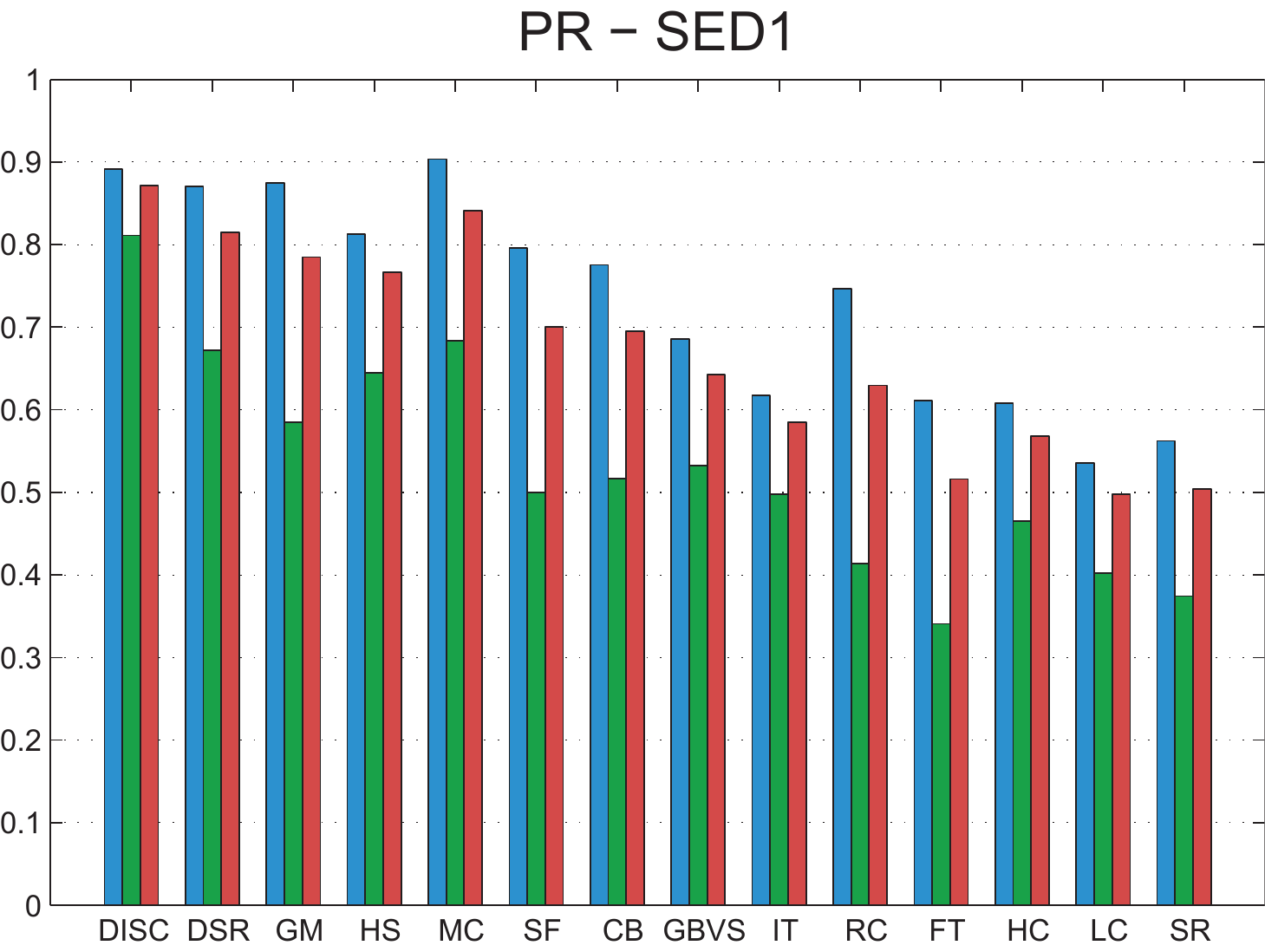}}
{\includegraphics[width=0.32\linewidth]{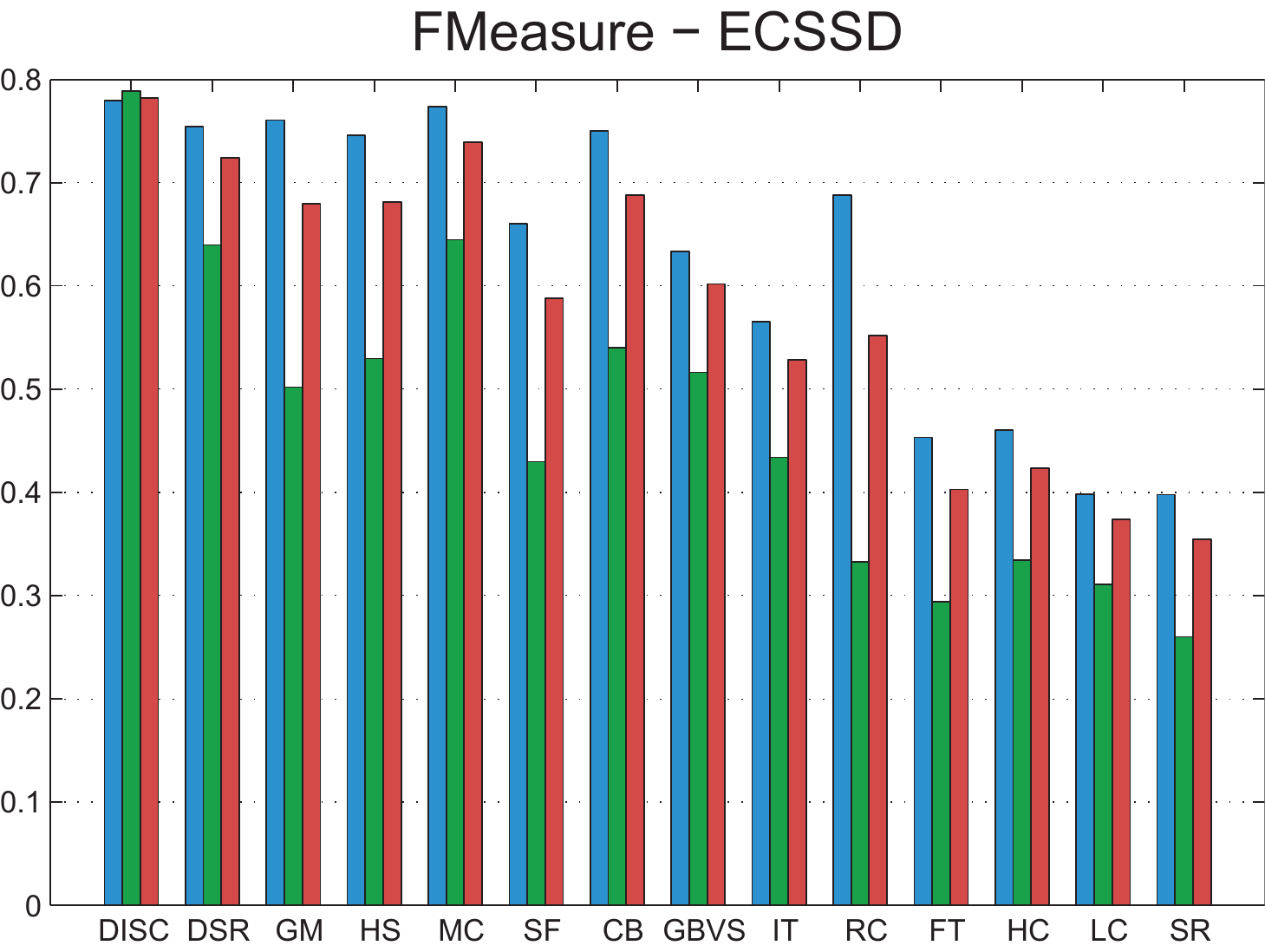}}
{\includegraphics[width=0.32\linewidth]{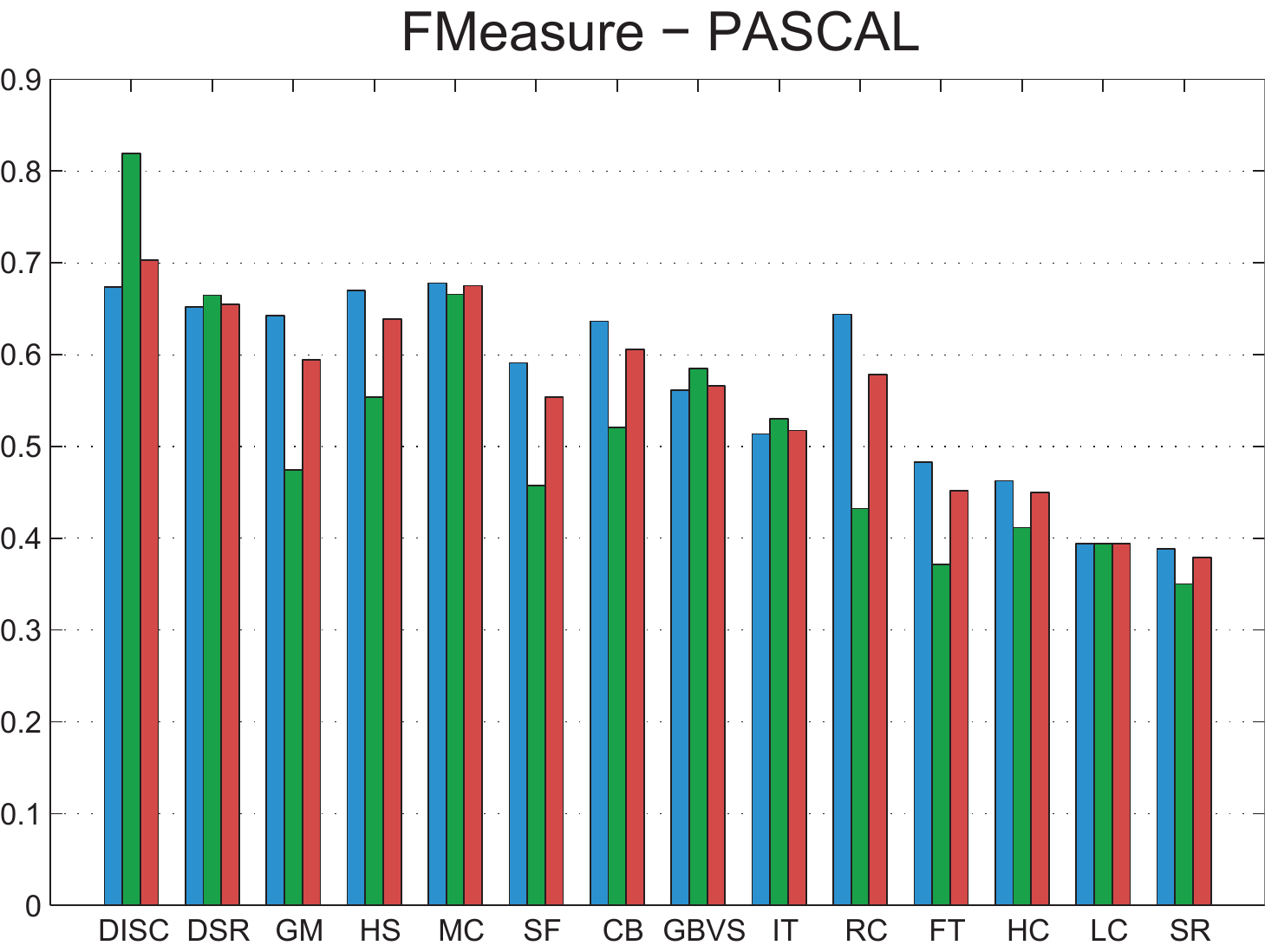}}
\subfigure[]{
\label{fig:subfig2_1} 
\includegraphics[width=0.32\linewidth]{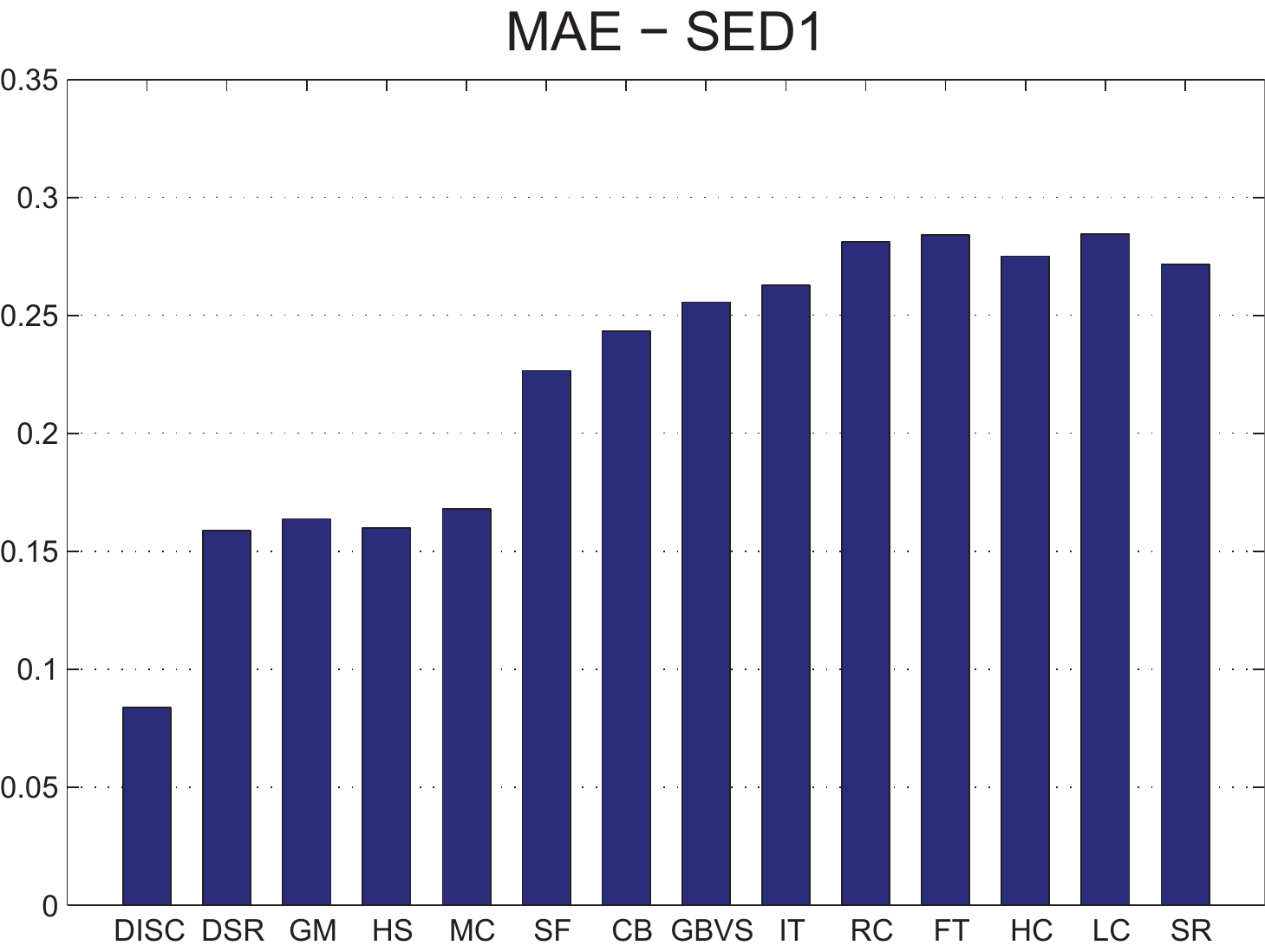}}
\subfigure[]{
\label{fig:subfig2_2} 
\includegraphics[width=0.32\linewidth]{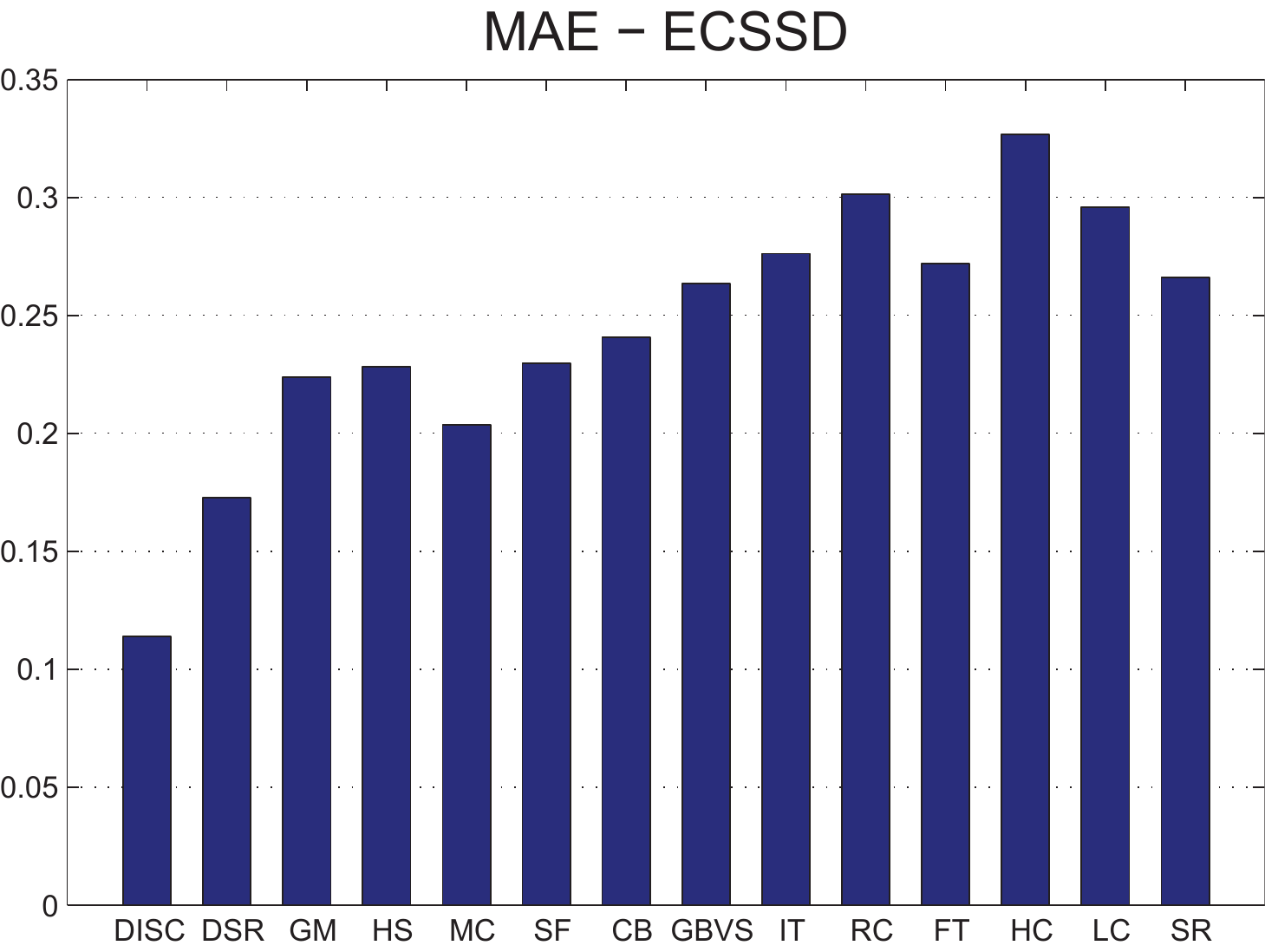}}
\subfigure[]{
\label{fig:subfig2_3} 
\includegraphics[width=0.32\linewidth]{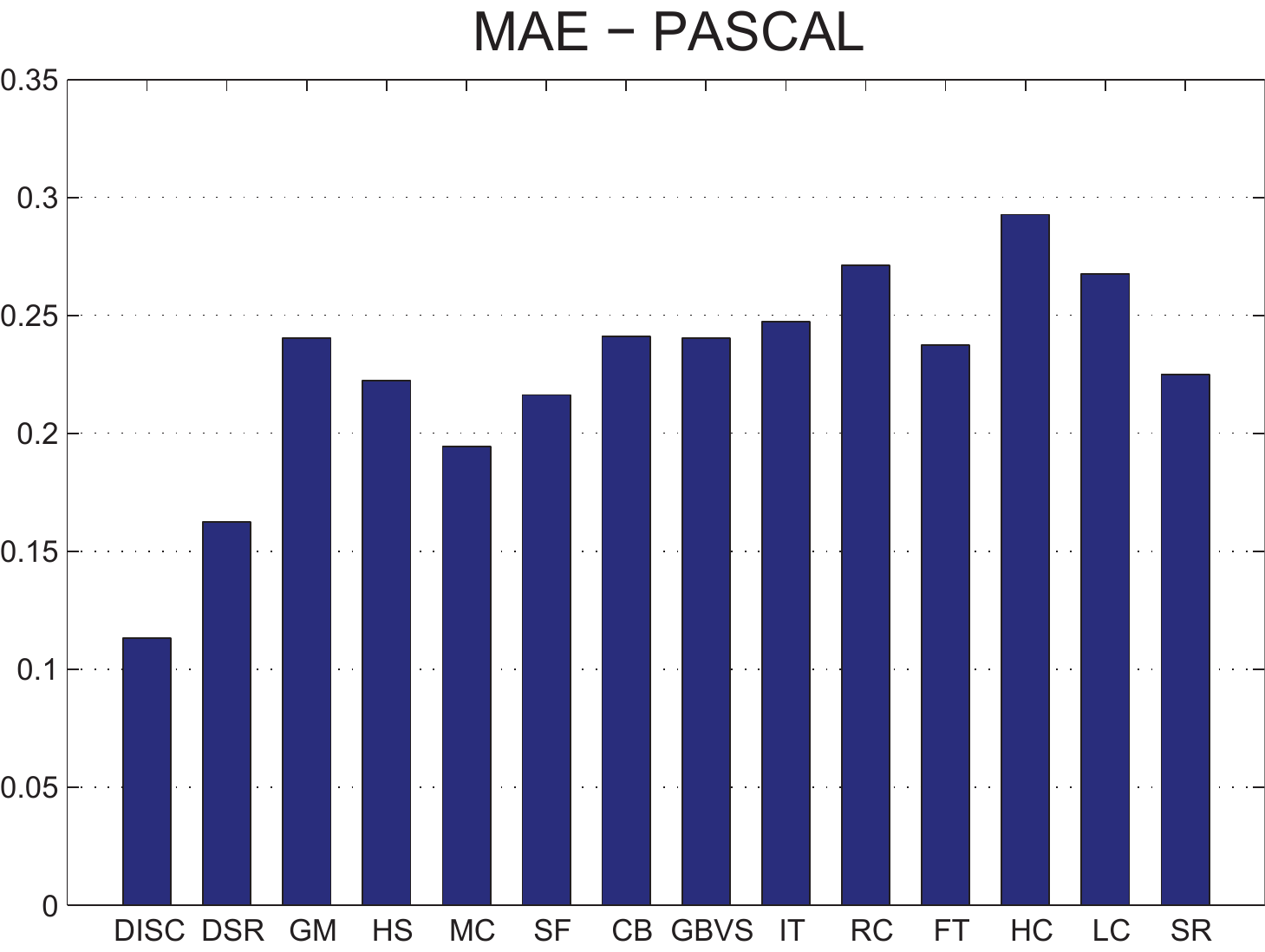}}
\caption{Experimental results on the (a) SED1, (b) ECSSD, and (c) PASCAL1500 datasets compared with previous works. Precision-recall curves (the first row), precision-recall bar with F-measure (the second row), and mean absolute error (the third row) show superior generalization ability of DISC framework. Note that our method still achieves state-of-the-art performance when it is learned on a small-scale set (i.e.,MSRA10K) without fine-tuning on the target datasets. Best viewed in color.}
\label{fig:result_gerneration}
\end{figure*}

\begin{figure*}
\centering
\includegraphics[width=0.92\linewidth]{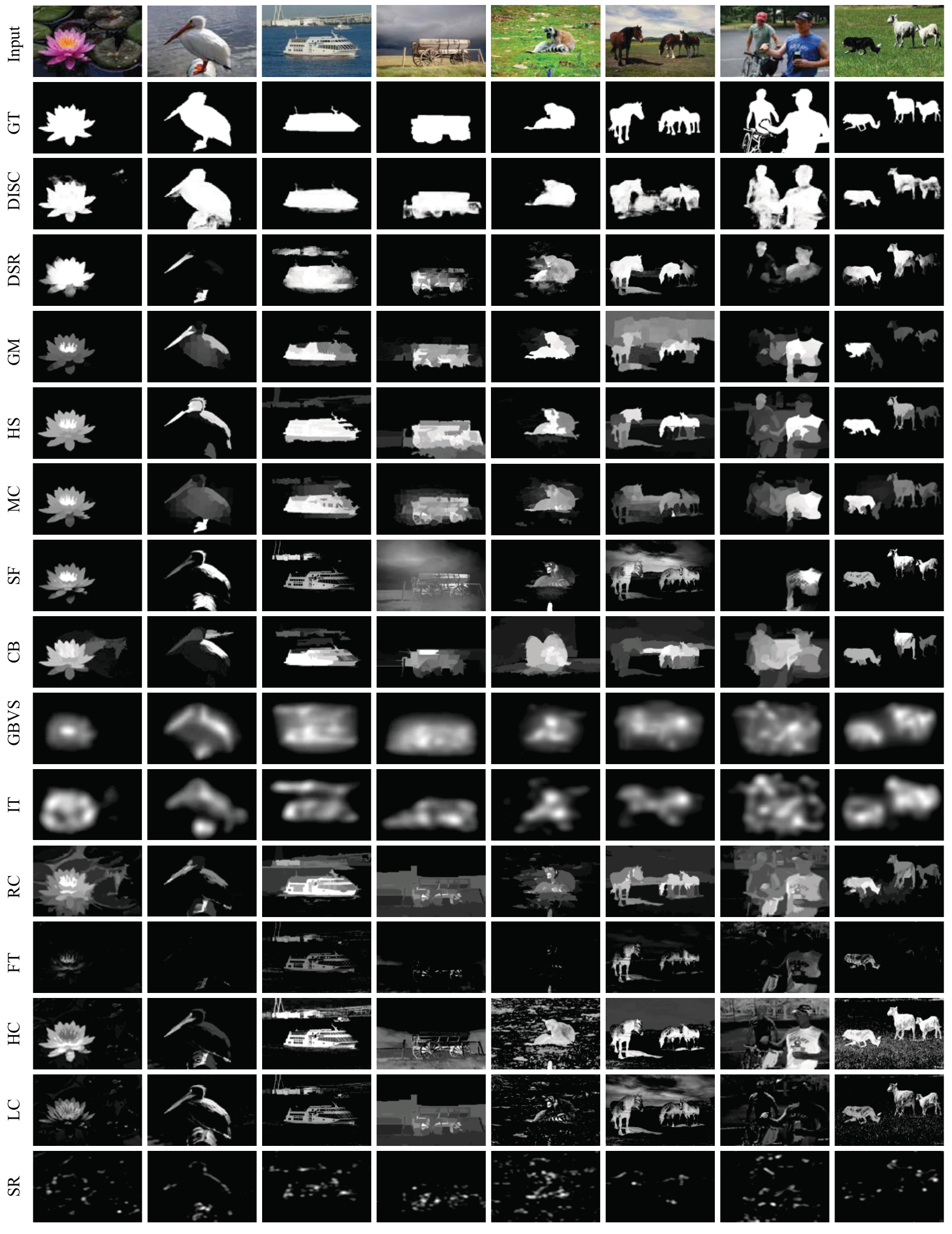}
\caption{Visual comparision with previous methods. The images are taken from MSRA10K (first two columns), SED1 (third and fourth columns), ECSSD (fifth and sixth columns), and PASCAL1500 (last two columms). Our DISC model not only highlights the overall objects but preserves boundary and structure details.}
\label{fig:all_visual_result}
\end{figure*}

\noindent \textbf{Evaluation protocol.}
We use Precision-Recall (PR) curves, $F_{0.3}$ metric, and Mean Absolute Error (MAE) to evaluate all the methods. Precision is the fraction of correct salient pixel assigned number in relation to all the detected salient pixel number while recall is the fraction of correct salient pixel number in relation to the ground truth number. Varying the threshold of saliency object segmentation from 0 to 255, we can plot the PR-curve.

$F_{0.3}$ metric applies an image adaptive threshold proposed by \cite{achanta2009frequency}. The adaptive threshold is defined as twice the mean saliency of the image

\begin{equation}
{T_f} = \frac{2}{{W \times H}}\sum\limits_{x = 1}^W {\sum\limits_{y = 1}^H {S(x,y)} },
\end{equation}
where $W$ and $H$ denote the width and height of the image respectively, and $S(x,y)$ denotes saliency value of the pixel at position $(x,y)$. The average precise and recall are obtained with the adaptive threshold above, and the F-measure is define as:

\begin{equation}
{F_{{\beta ^2}}} = \frac{{(1 + {\beta ^2})Precision \times Recall}}{{{\beta ^2} \times Precision + Recall}}.
\end{equation}

The same as \cite{achanta2009frequency}, we set $\beta^{2}=0.3$ to weigh precision more than recall.

As indicated in \cite{perazzi2012saliency}, PR curves and $F_{0.3}$ metric provide a quantitive evaluation, while MAE provides a better estimate of the dissimilarity between the continuous saliency map and binary ground truth, which is defined as below

\begin{equation}
MAE = \frac{1}{{W \times H}}\sum\limits_{x = 1}^W {\sum\limits_{y = 1}^H {|S(x,y)} }  - GT(x,y)|.
\end{equation}

\noindent \textbf{Implementation details.}
For the balance between computational efficiency and accuracy, we resize each input image to $256\times256$ for both CNNs and the output size of the fist CNN is $64\times64$ while that of the second CNN is $128\times128$. We implement the two CNNs under the Caffe framework \cite{jia2014caffe}, and train them using stochastic gradient descent (SGD) with momentum of 0.9, and weight decay of 0.0005. The learning rate for training the first CNN is initialized as $10^{-6}$ with a batch size of 32 and that for training the second CNN is initialized as $10^{-7}$ with a batch size of 2. We train the first CNN for about 90 epochs and the second CNN for roughly 55 epochs, and the training procedure costs nearly two days in all. During inference, we first assign -1 and 1 for the pixels with saliency scores smaller than -1 and larger than 1, respectively, and then transform them to [0, 255] via a simple linear normalization. Our method can calculate a $128\times128$ normalized saliency map within about 75ms on a single NVIDIA GeForce GTX TITAN Black. It is relatively time-saving compared to previous state-of-the-art approaches. For example, Sparse Reconstruction \cite{li2013saliency} costs about 3.536s and Hierarchical Saliency \cite{yan2013hierarchical} runs in about 397ms on a desktop with an Intel i7 3.4GHz CPU and 8GB RAM.

\vspace{-6pt}
\subsection{Comparison with State-of-the-art Methods}
In this subsection, we evaluate the proposed method on MSRA10K \cite{cheng2011global} dataset. In our experiment, we randomly divide MSRA10K into two subsets, one subset of 9,000 images for training and the other subset of 1,000 images for verification. We repeat the experiment for three times, and report the average result.
We compare DISC with thirteen recent state-of-the-art approaches: Context-Based saliency (CB) \cite{jiang2011automatic}, Sparse Reconstruction (DSR) \cite{li2013saliency}, Graph-based Manifold ranking (GM) \cite{yang2013saliency}, Hierarchical Saliency (HS) \cite{yan2013hierarchical}, Markov Chain saliency (MC) \cite{jiang2013saliency}, Saliency Filter (SF) \cite{perazzi2012saliency}, Visual attention mesure (IT) \cite{itti1998model}, Graph-Based Visual Saliency (GBVS) \cite{harel2006graph}, Frequency-Tuned saliency (FT) \cite{achanta2009frequency}, Spatial-temporal Cues (LC) \cite{zhai2006visual}, Histogram-based Contrast (HC) \cite{cheng2011global}, and Region-based Contrast (RC) \cite{cheng2011global}. We adopt the implementations from the original authors to generate the results for CB, DSR, SM, HS, MC, SF and use the codes provided by \cite{cheng2011global} to generate the results of LC, FT, HC, RC, SR. The results of IT, GBVS are produced using the codes provided by \cite{harel2006graph}. We normalize all the saliency scores to [0, 255].

The results of PR curve, $F_{0.3}$ metric, MAE on MSRA10K dataset are shown in Figure \ref{fig:result_MSRA} respectively. Based on the PR-curve, although some previous methods, such as DSR, GM, HS, MC, have achieved more than 93\% accuracy, DISC still has significant improvement over all of them, reaching 97.3\%. On the other side, the minimal recall value of DISC is 34\%, significantly higher than those of the other methods, because the saliency maps computed by DISC contain more salient pixels with the saliency scores of 255. Based on $F_{0.3}$, our method performs consistently better than the others, as the precision is comparable with other state-of-the-art result but the recall is 15.3\% higher than the best previous work. Besides, the MAE of DISC is significantly lower than the others, which suggests we preserve the details much better.

Some saliency maps produced by DISC and previous works are depicted in Figure \ref{fig:all_visual_result} for visual comparison. It can be seen that our method not only highlights the overall salient objects, but also preserves the detail very well.

These comparisons suggest that DISC outperforms other state-of-the-art algorithms by large margins. The main reason for the superior performance is our coarse-to-fine deep architecture which is capable of capturing different level of image saliency information. More specifically, compared with traditional saliency models using hand-craft features, the deep models enable learning very powerful features by incorporating domain knowledge (i.e., how to define and model the image saliency) into neural networks and leveraging the large-scale data for effective learning. Actually, we believe that is very common reason for the success of deep learning in several vision tasks.


\begin{figure}[htp]
\centering
\subfigure[]{
\label{fig:pr_dl} 
\includegraphics[width=0.45\linewidth]{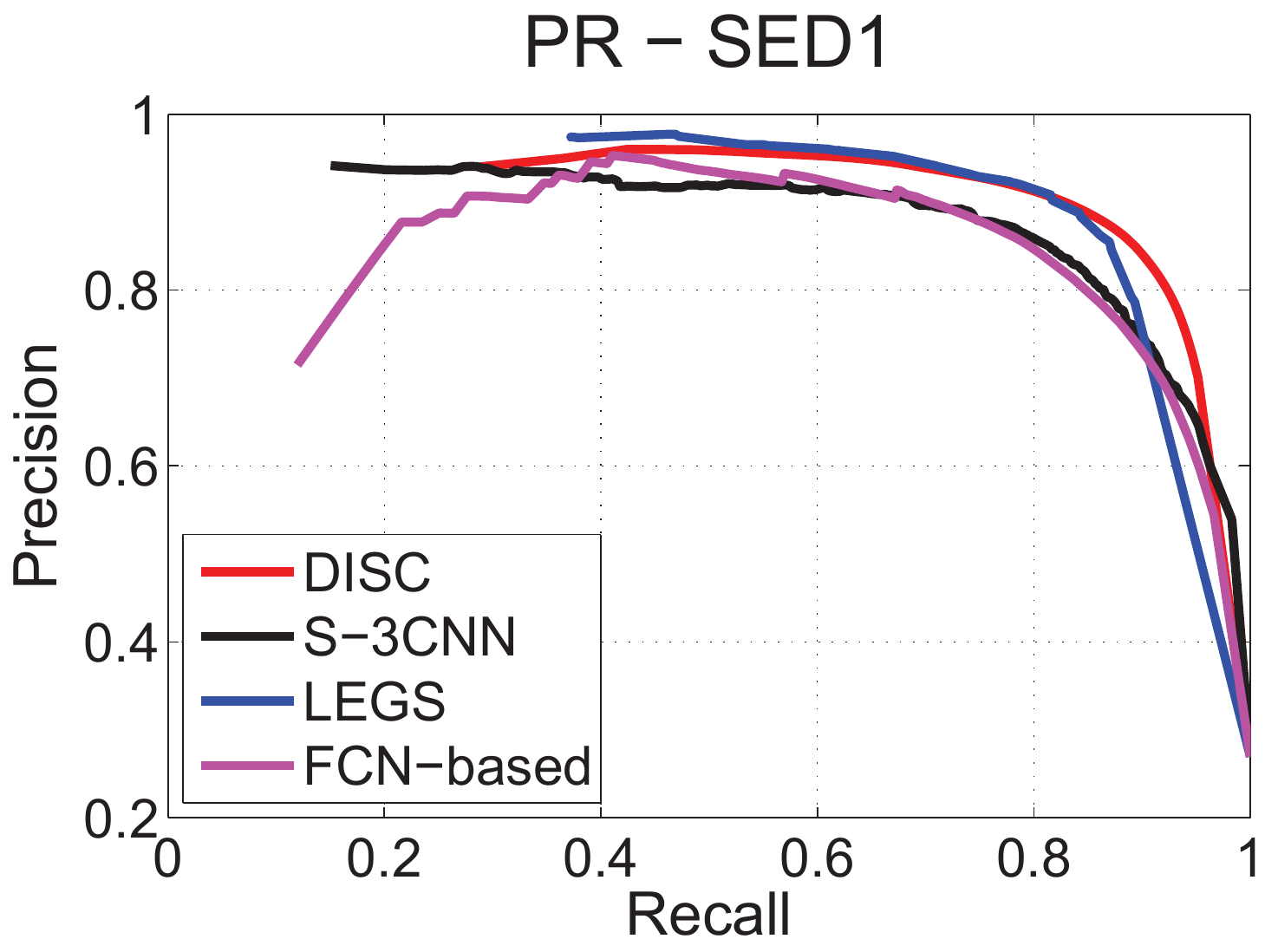}}
\subfigure[]{
\label{fig:result_task-oriented} 
\includegraphics[width=0.45\linewidth]{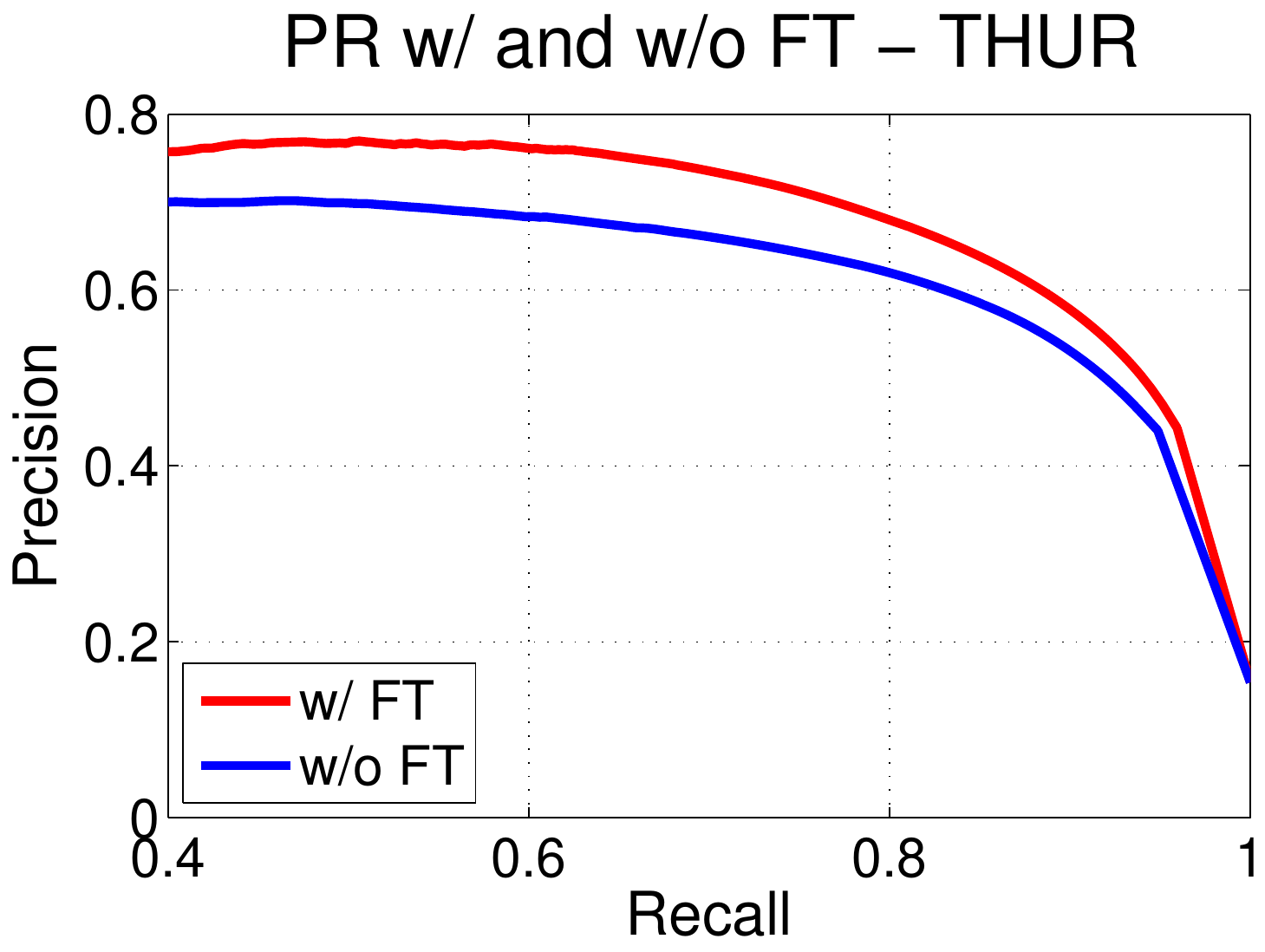}}
\vspace{-6pt}
\caption{(a) Precision-recall curve of DISC and other CNN-based methods on SED1 dataset. (b) Precision-recall curve of DISC with and without fine tuning on THUR dataset. Best viewed in color.}
\label{fig:dl_task}
\vspace{-20pt}
\end{figure}

\vspace{-6pt}
\subsection{Comparision with CNN-based Methods}
To further demonstrate the effectiveness of the proposed architecture, we provide the results of comparison between DISC and other CNN-based methods: S-3CNN \cite{li2015visual}, LEGS \cite{wang2015deep} and a FCN-based method. S-3CNN and LEGS are two recent-public works which utilize multi-scale CNN structure for salient object detection. FCN \cite{long2015fully} takes input of arbitrary size and produces a correspondingly-sized dense label map, and it has shown convincing results for pixel-wise labeling such as semantic image segmentation, thus we also apply the FCN to salient object detection for comparison. We adopt the trained models and source codes provided by the original authors to generate the results for S-3CNN and LEGS. For FCN-based method, we utilize the best FCN-VGG16-8s architecture and train it using 9,000 images of MSRA-10K dataset. Since the training datasets are quite different, we test all the methods on SED1 dataset, which has no overlap with all the datasets used for models training, for fair cross-dataset comparison. The experiments are all carried out on a single NVIDIA GeForce GTX TITAN Black and the results are shown in Figure \ref{fig:pr_dl} and Table \ref{table:deep learning}. DISC performs consistently better than other CNN-based methods and the running time is 75ms per image, dramatically faster than S-3CNN and LEGS while slightly faster than FCN-based method. It demonstrates the effectiveness and efficiency of our coarse-to-fine architecture.

\begin{table}[htp]
\centering
\begin{tabular}{c|c|c|c|c|c}
\hline
& Precision & Recall & F-measure & MAE & Time (ms) \\
\hline\hline
DISC &  0.892 & 0.811 & 0.872 & 0.084 & 75 \\
\hline
S-3CNN & 0.852 & 0.746 & 0.825 & 0.132 & 17700 \\
\hline
LEGS & 0.912 & 0.740 & 0.865 & 0.107 & 2330 \\
\hline
FCN-based & 0.873 & 0.730 & 0.836 & 0.129 & 90 \\
\hline
\end{tabular}
\vspace{2pt}
\caption{The precision-recall with F-measure, MAE, and running time of DISC and other three CNN-based methods on SED1 dataset.}
\vspace{-20pt}
\label{table:deep learning}
\end{table}

\vspace{-6pt}
\subsection{Performance of Generalization}
In this subsection, we evaluate the generalization performance of DISC. It is a labor-intensive and time-consuming job to collet enough labeled data to learn particular model for each scenario. Therefore, transferring a learnt model to current scenario without significant degradation is a more practical methods. To assess how well our model generalized to other datasets, we evaluate DISC on three datasets, i.e., SED1, ECSSD, and PASCAL1500. As discuss in experiment setting, the images of three datasets are collected in three different scenarios. We directly test the performance on these three datasets with the model learnt on MSRA-10K. The results are shown in Figure \ref{fig:result_gerneration}. Although the model is trained on the other dataset, it outperforms all other previous methods based on the three evaluation metrics. It is amazing because the three datasets are significantly different both in salient objects and background, which demonstrate the excellent generalization ability of DISC.

\begin{figure}[htp]
\centering
\subfigure[]{
\label{fig:finetuning_1} 
\includegraphics[width=0.45\linewidth]{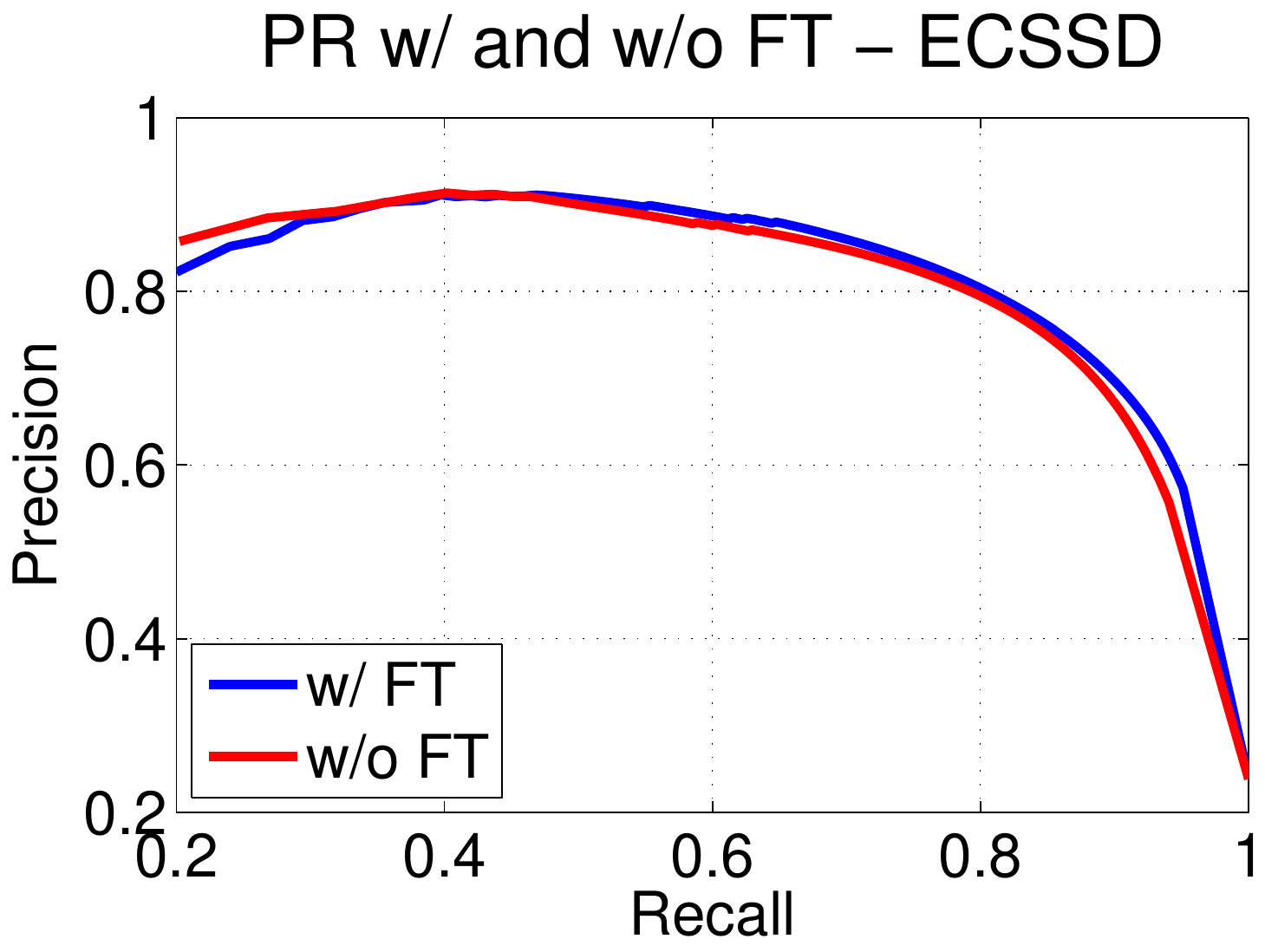}}
\subfigure[]{
\label{fig:finetuning_2} 
\includegraphics[width=0.45\linewidth]{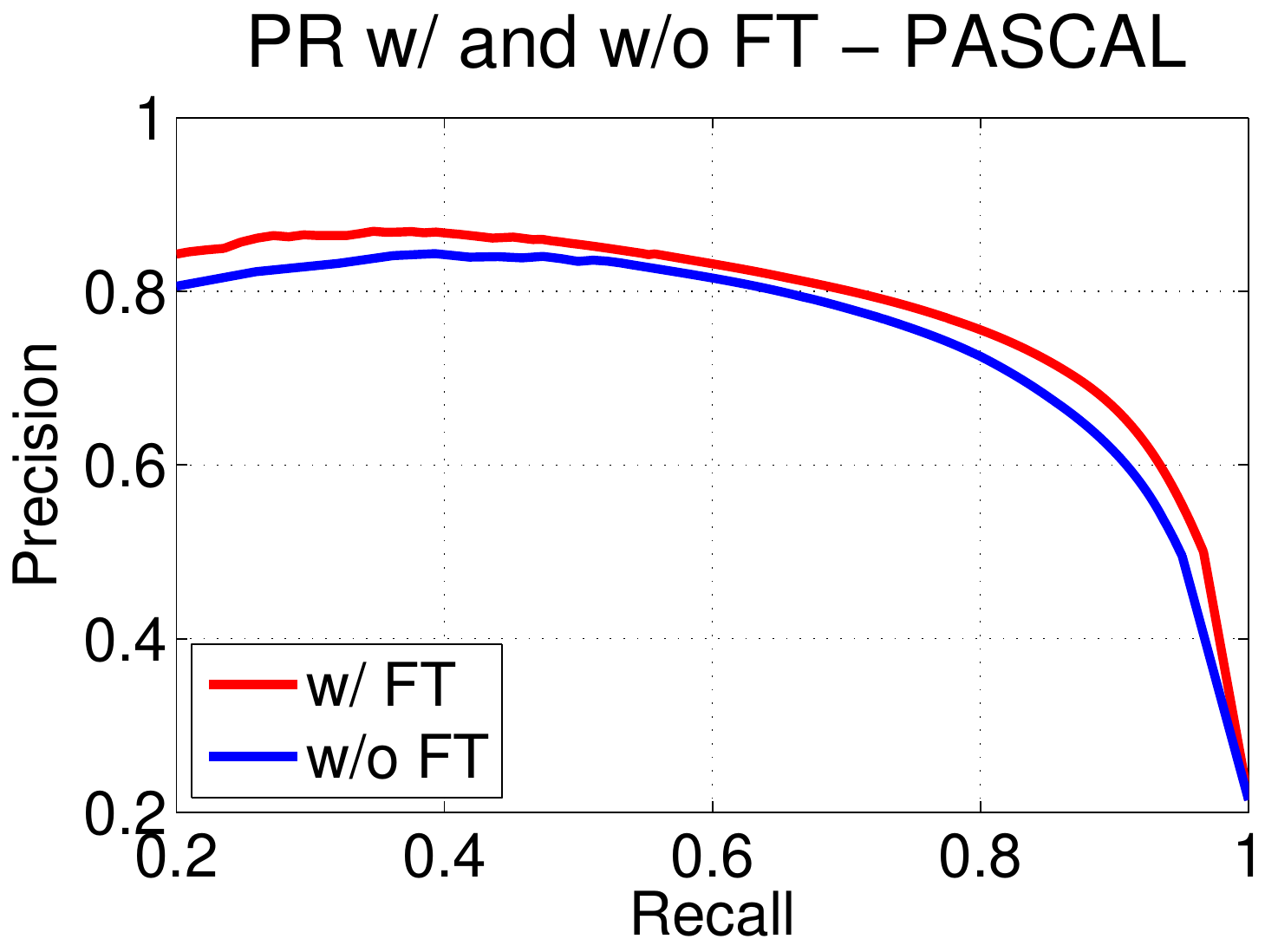}}
\vspace{-6pt}
\caption{Experimental results of fine tuning on the ECSSD and PASCAL1500 datasets. Precision-recall curve of our DISC framework with and without fine tuning (FT) on (a) ECSSD and (b) PASCAL1500. Best viewed in color.}
\vspace{-12pt}
\label{fig:finetuning}
\end{figure}

\begin{figure*}[!t]
\centering
\subfigure[]{
\label{fig:coarse_to_fine}}
\includegraphics[width=0.23\linewidth]{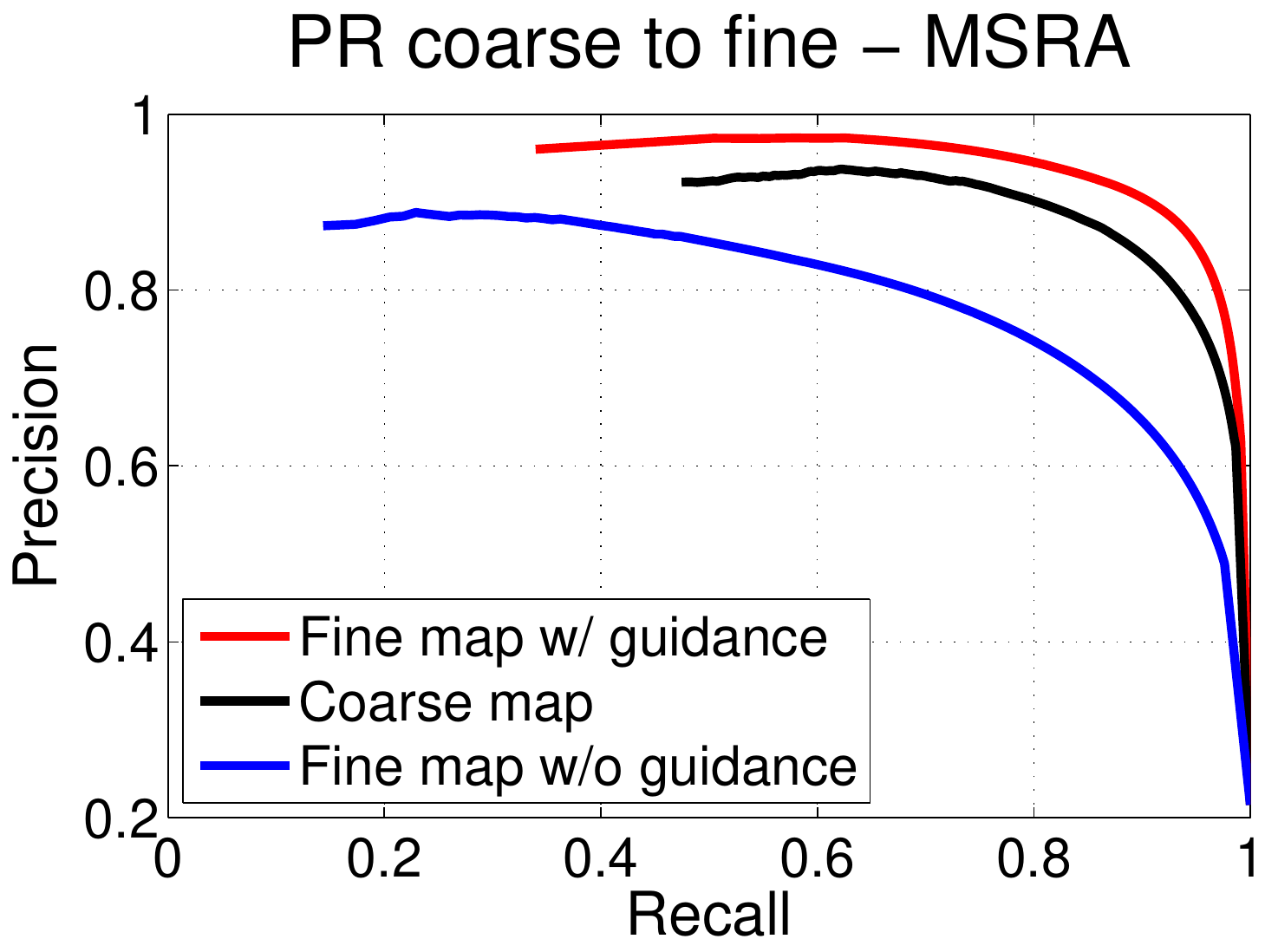}
\subfigure[]{
\label{fig:pr_slci}}
\includegraphics[width=0.23\linewidth]{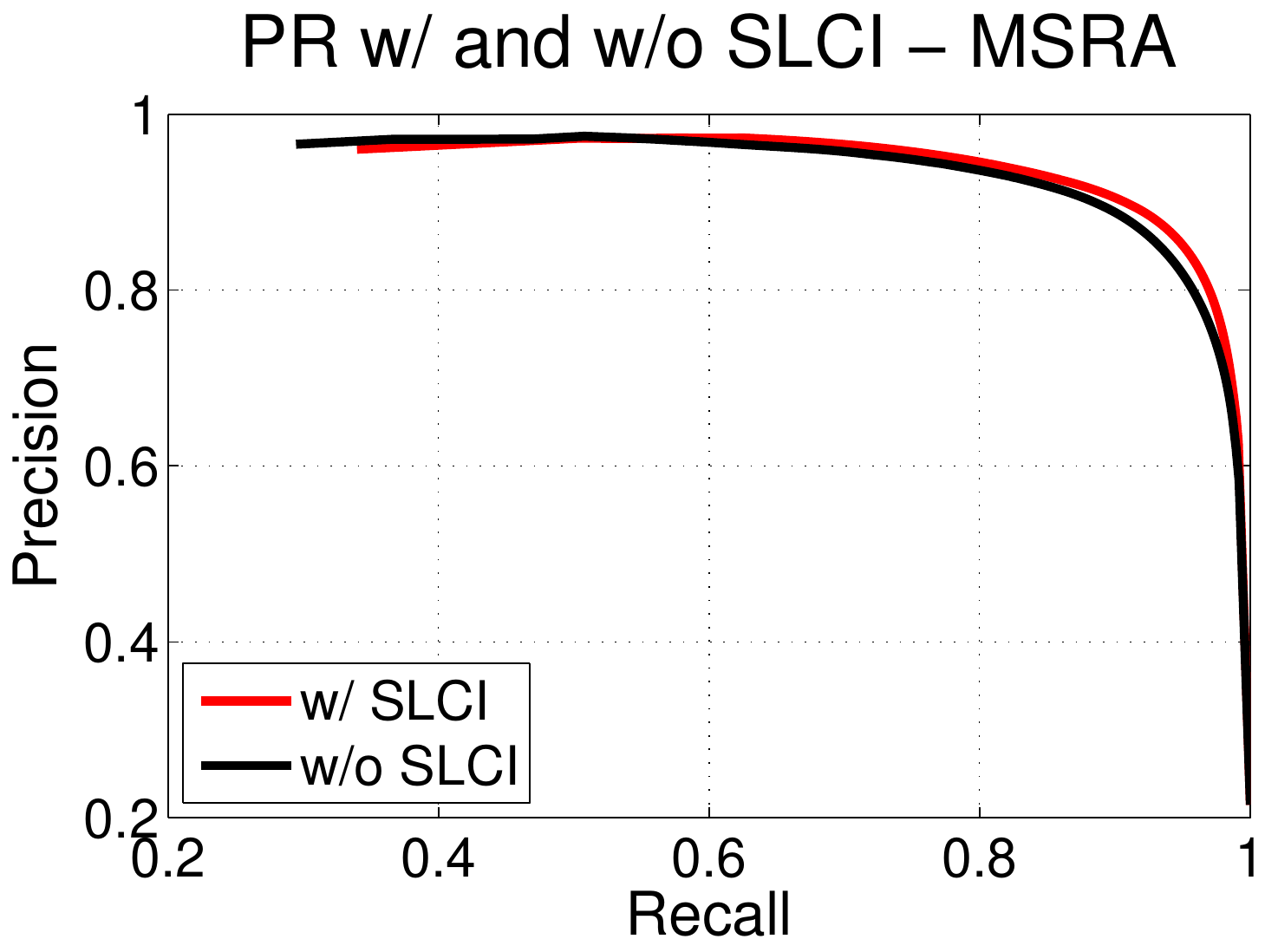}
\subfigure[]{
\label{fig:spatial_regularization}}
\includegraphics[width=0.23\linewidth]{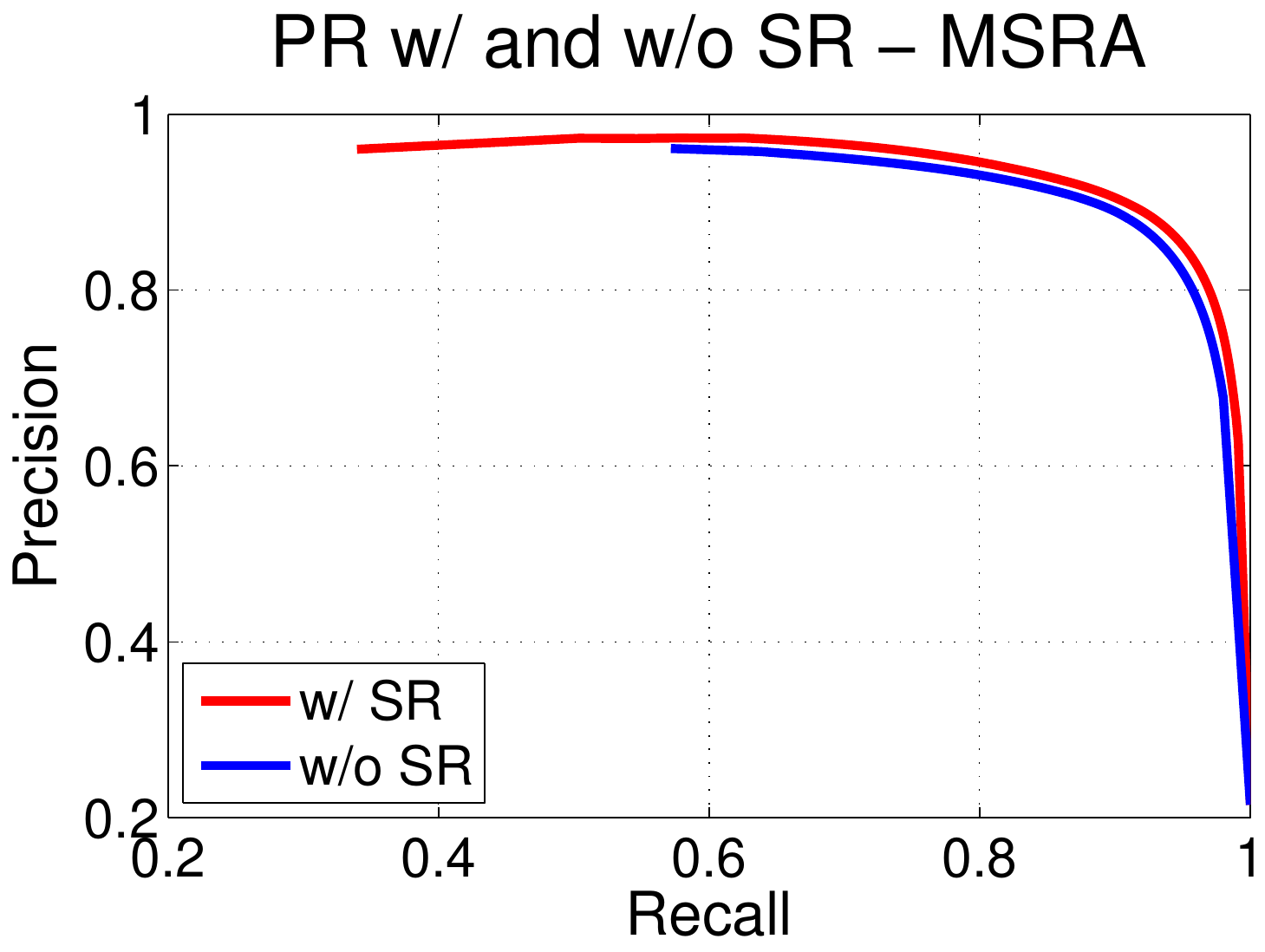}
\subfigure[]{
\label{fig:result_hinge_sigmoid}}
\includegraphics[width=0.23\linewidth]{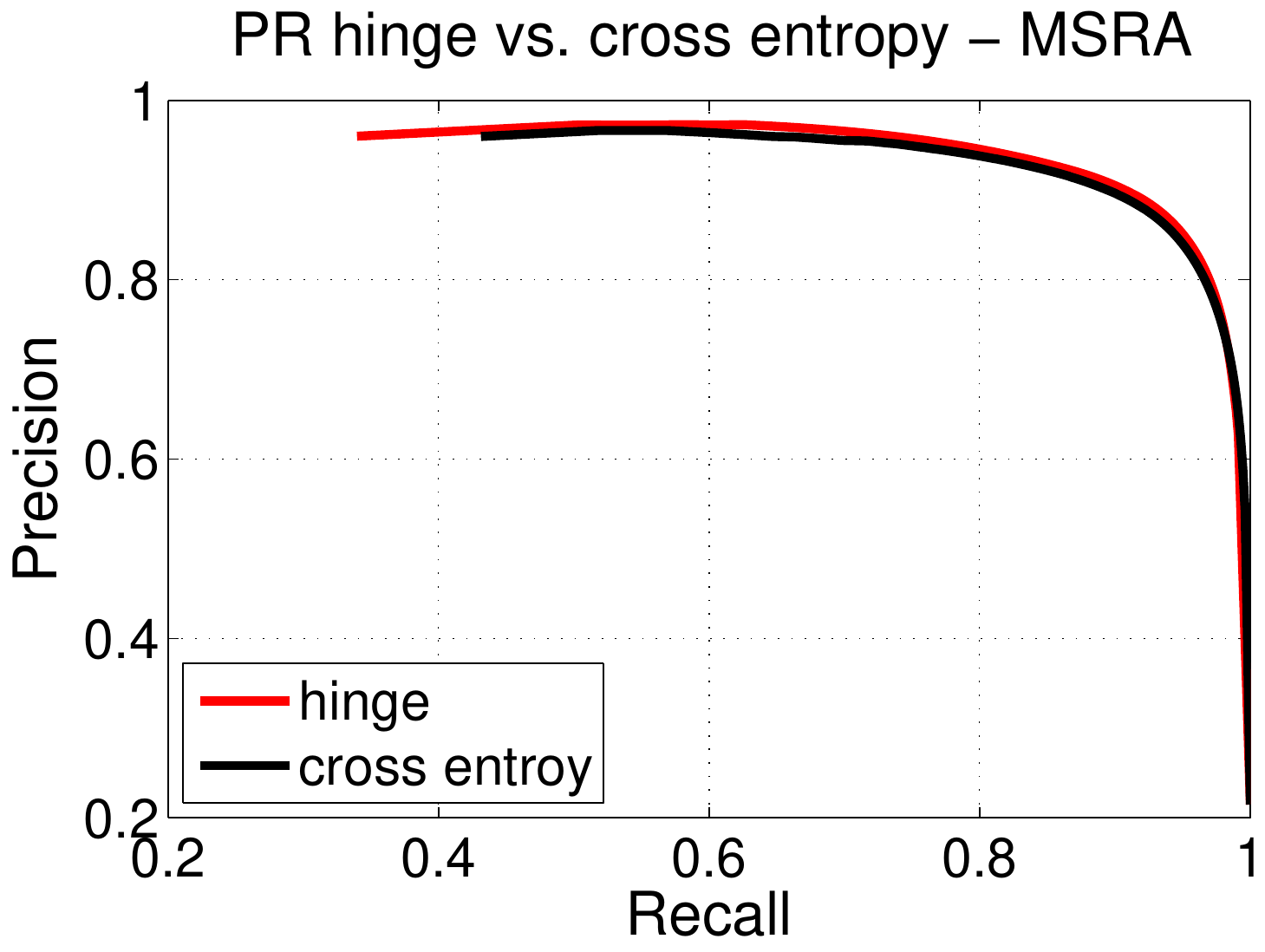}
\vspace{-6pt}
\caption{(a) Precision-recall curve of coarse maps and fine maps (with and without guidance) on the MSRA-10k dataset. (b) Precision-recall curve of our DISC with and without superpixel-based local context information (SLCI) on the MSRA-10k dataset. (c) Precision-recall curve with and without spatial regularization on the MSRA-10k dataset. (d) Precision-recall curve of our method trained with hinge loss vs. cross entropy loss on the MSRA-10k dataset. Best viewed in color.}
\vspace{-20pt}
\label{fig:result_sub}
\end{figure*}

\vspace{-10pt}
\subsection{Fine Tuning on Other Datasets}
Although the result is quite outstanding on the ECSSD and PASCAL1500 datasets without fine tuning, we retrain on these two datasets to get better result. We do not retrain our model on the SED1 dataset, because it contains only 100 images and is too small for retraining. We randomly select 600 images from ECSSD and 1000 images from PASCAL1500 for fine tuning respectively and the rest is for testing. As Figure \ref{fig:finetuning} and Table \ref{table:finetuning} show, the result after fine tuning is slightly better than before. There are two factors that lead to such phenomenon.
The first is the generalization ability of DISC is excellent and it can deal with variable situations as discussed before. The second is that the samples for fine tuning may be insufficient.

\begin{table}[!t]
\centering
\begin{tabular}{c|c|c|c|c}
\hline
\multirow{2}{*}{Dataset} &
\multicolumn{2}{c|}{F-measure} &
\multicolumn{2}{c}{MAE} \\
\cline{2-5}
& w/ ft  & w/o ft & w/ ft  & w/o ft \\
\hline\hline
ECSSD & 0.779 & 0.784 & 0.117 & 0.114 \\
\hline
PASCAL & 0.738 & 0.712 & 0.116 & 0.114 \\
\hline
\end{tabular}
\vspace{2pt}
\caption{The F-measure and mean absolute error with and without fine tuning on the ECSSD and PASCAL datasets.}
\vspace{-30pt}
\label{table:finetuning}
\end{table}

\vspace{-10pt}
\subsection{Task-oriented Adaptation}

We select the 6,233 images with pixel-accurate ground truth annotations to evaluate the performance our model with task-oriented adaptation. We divide the 6,223 images into two parts, 5,233 images for fine tuning and 1,000 images for testing. We first produce the saliency maps for the 1,000 testing images with the model learned on the MSRA10K dataset. Then we fine tune DISC with the 5,233 training images. The PR curves with and without fine tuning are shown in Figure \ref{fig:result_task-oriented} while the corresponding precision-recall with F-measure and MAE are listed in Table \ref{table:task-oriented}. Since the ground truth only labels the specific objects, but the model without retaining highlights all the salient objects, the precision of result without fine tuning is very low. After fine tuning, the precision improves significantly while the recall keeps nearly the same. It suggests that the pixel number we mislabel as saliency decreases. In the other word, we do only highlight the specific classes of objects, ignoring others.


\begin{table}[htp]
\centering
\begin{tabular}{c|c|c|c|c}
\hline
& Precision & Recall & F-measure & MAE \\
\hline\hline
w/ FT & 0.664 & 0.810 & 0.693 & 0.084 \\
\hline
w/o FT & 0.586 & 0.817 & 0.626 & 0.118 \\
\hline
\end{tabular}
\vspace{2pt}
\caption{The precision-recall with F-measure and mean absolute error with and without fine tuning on the THUR dataset.}
\vspace{-30pt}
\label{table:task-oriented}
\end{table}

\subsection{Evaluation and Analysis}

In this part, we would like to analyze and discuss each component of the proposed model so as to evaluate the actual contribution of corresponding component.

\subsubsection{Contribution of progressive representation learning}

\begin{figure}[htp]
\centering
\includegraphics[width=0.9\linewidth]{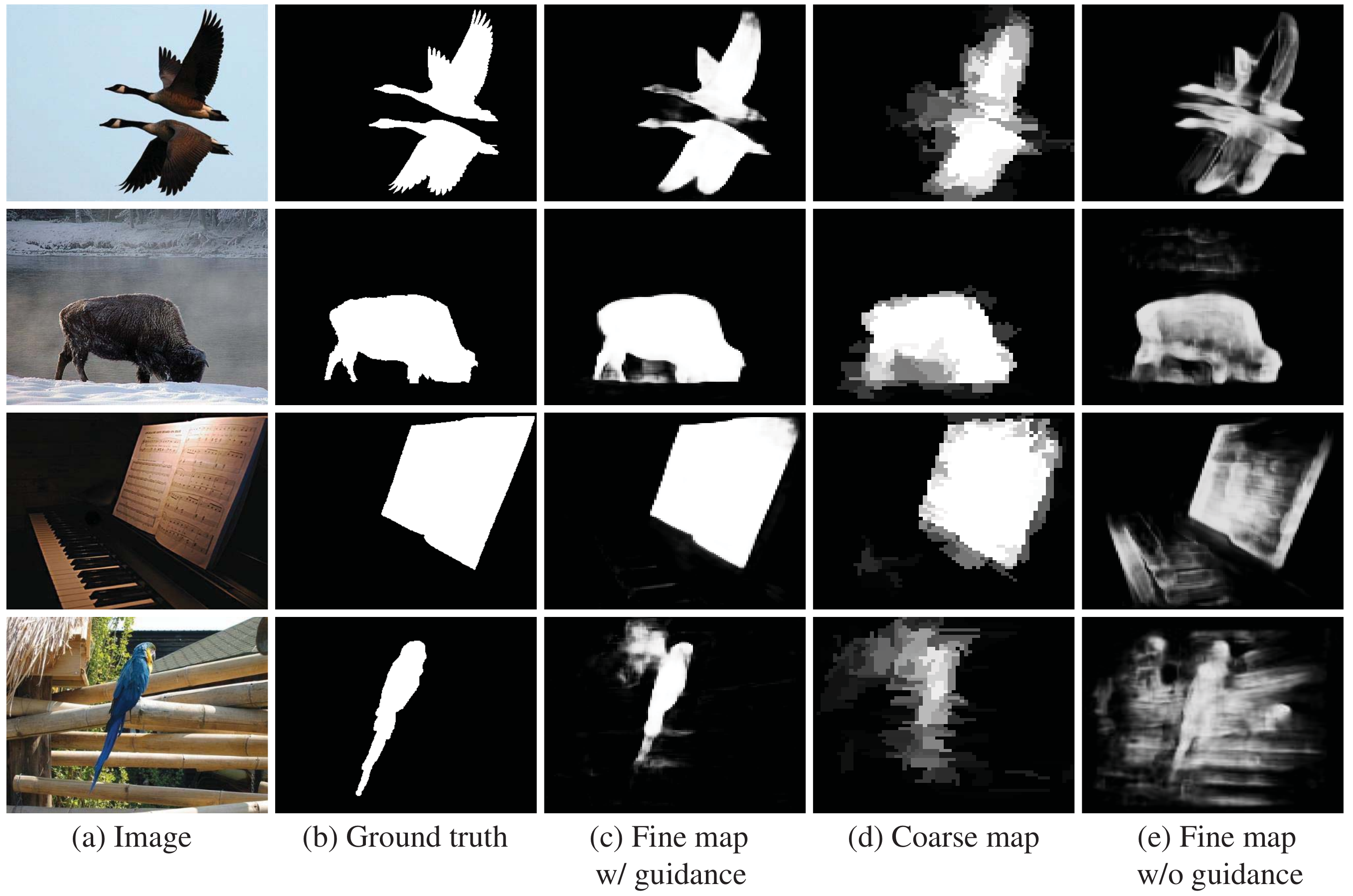}
\vspace{-12pt}
\caption{Visual comparison of coarse maps and fine maps (with and without guidance). The samples are taken from the MSRA and ECSSD datasets.}
\label{fig:progress_rep}
\end{figure}

The progressive representation learning framework is designed to learn the saliency representation in a coarse-fine manner, with the coarse representation capturing the object location and global structural information, while the fine representation further refining the object details. We evaluate the contribution of our progressive representation learning framework by comparing its performance with those when using only coarse representation, and using the fine maps without guidance. We show some examples of these three methods in Figure \ref{fig:progress_rep}. It is clear that the coarse maps are capable of highlighting the locations and shapes of the salient objects roughly, but the details, especially object boundaries and subtle structures, are easily lost. Overall, these coarse maps are well suited to help learn the fine representation. In contrast, generating the fine maps without guidance usually has two main drawbacks. First, it may miss the interior contents while emphasizing the boundaries of salient objects if the background is relatively clean (e.g., first two examples in Figure \ref{fig:progress_rep}). Second, it is likely to mistakenly highlight some of the background regions particularly when the background is cluttered (e.g., the last two examples in Figure \ref{fig:progress_rep}). The proposed DISC framework combines the advantage of coarse and fine maps, producing more accurate and structure-preserved results. We also present the quantitative comparisons in Figure \ref{fig:coarse_to_fine} and Table \ref{table:ctf}. They show that the performance of our results dramatically suppresses the results of coarse maps and fine maps without guidance. This comparison well demonstrates the effectiveness of progressive representation learning.

\vspace{-6pt}
\begin{table}[htp]
\centering
\begin{tabular}{c|c|c|c|c}
\hline
& Precision & Recall & F-measure & MAE \\
\hline\hline
Fine map w/ guidance & 0.872 & 0.927 & 0.884 & 0.052 \\
\hline
Coarse map & 0.819 & 0.900 & 0.837 & 0.076 \\
\hline
Fine map w/o guidance & 0.744 & 0.785 & 0.753 & 0.143 \\
\hline
\end{tabular}
\vspace{2pt}
\caption{The precision-recall with F-measure and mean absolute error coarse maps and fine maps (with and without guidance) on the MSRA dataset.}
\vspace{-30pt}
\label{table:ctf}
\end{table}

\subsubsection{Contribution of superpixel-based local context information (SLCI)}
\begin{figure}[htp]
\centering
\includegraphics[width=0.98\linewidth]{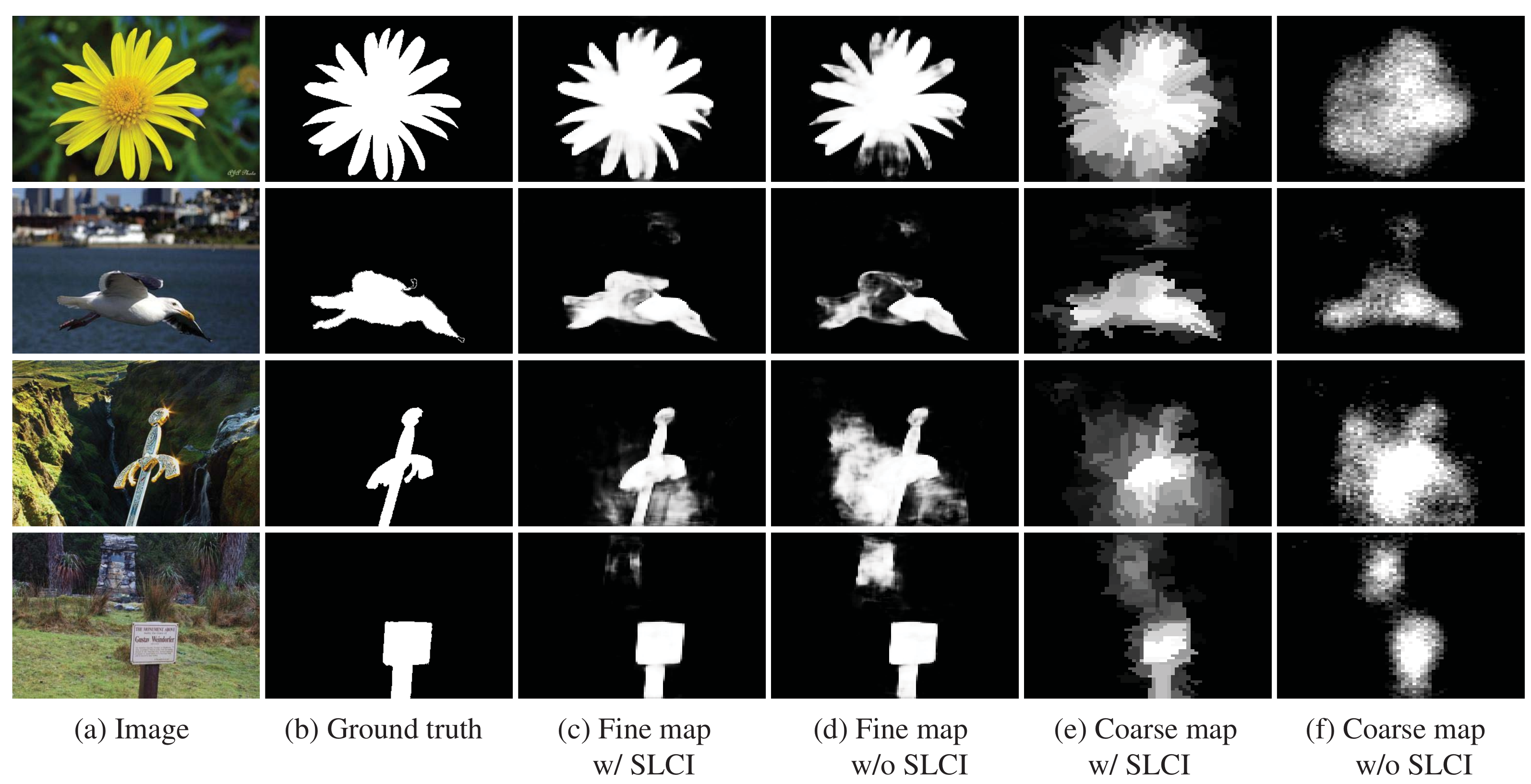}
\vspace{-6pt}
\caption{Visual comparison of coarse/fine maps with and without SLCI. The samples are taken from the MSRA10K and ECSSD datasets.}
\label{fig:SLCI}
\end{figure}

In this part, we analyze the contribution of SLCI. Here, we simply remove the ISS and ISV pooling layers from the first CNN, and then retrain two networks. Some saliency maps generated with and without SLCI are shown in Figure \ref{fig:SLCI}. We first compare the results of the coarse maps. It can be seen that the removal of SLCI leads to poor structure preserved performance. This is not surprising because the first CNN has to learn the coarse saliency representation based solely on the visual cues while neglecting local structural constraints. As a result, it inevitably confuses some small regions from salient objects or background that appear similar, and also produces counterintuitive structures or shapes for the target objects (see Figure \ref{fig:SLCI}(f)). Aided by the SLCI, the local structural information can be better preserved.
As discussed in the previous part, the fine map depends heavily on the quality of the coarse map. Thus, the corresponding fine maps also suffer from similar problems above. We propose to embed the SLCI into the first CNN, which improves structure preserved performance, as shown in Figure \ref{fig:SLCI}(c) and (e).
Meanwhile, quantitative comparisons on three metrics are provided in Figure \ref{fig:pr_slci} and Table \ref{table:SLCI}, which show the performance is better than that without SLCI.

\begin{table}[htp]
\centering
\begin{tabular}{c|c|c|c|c}
\hline
& Precision & Recall & F-measure & MAE \\
\hline\hline
w/ SLCI  & 0.872 & 0.927 & 0.884  & 0.052 \\
\hline
w/o SLCI & 0.873 & 0.908 & 0.881 & 0.056 \\
\hline
\end{tabular}
\vspace{2pt}
\caption{The precision-recall with F-measure and mean absolute error with and without superpixel-based local context information (SLCI) on the MSRA dataset.}
\vspace{-30pt}
\label{table:SLCI}
\end{table}

\subsubsection{Contribution of spatial regularization (SR)}
\begin{figure}[htp]
\centering
\includegraphics[width=0.98\linewidth]{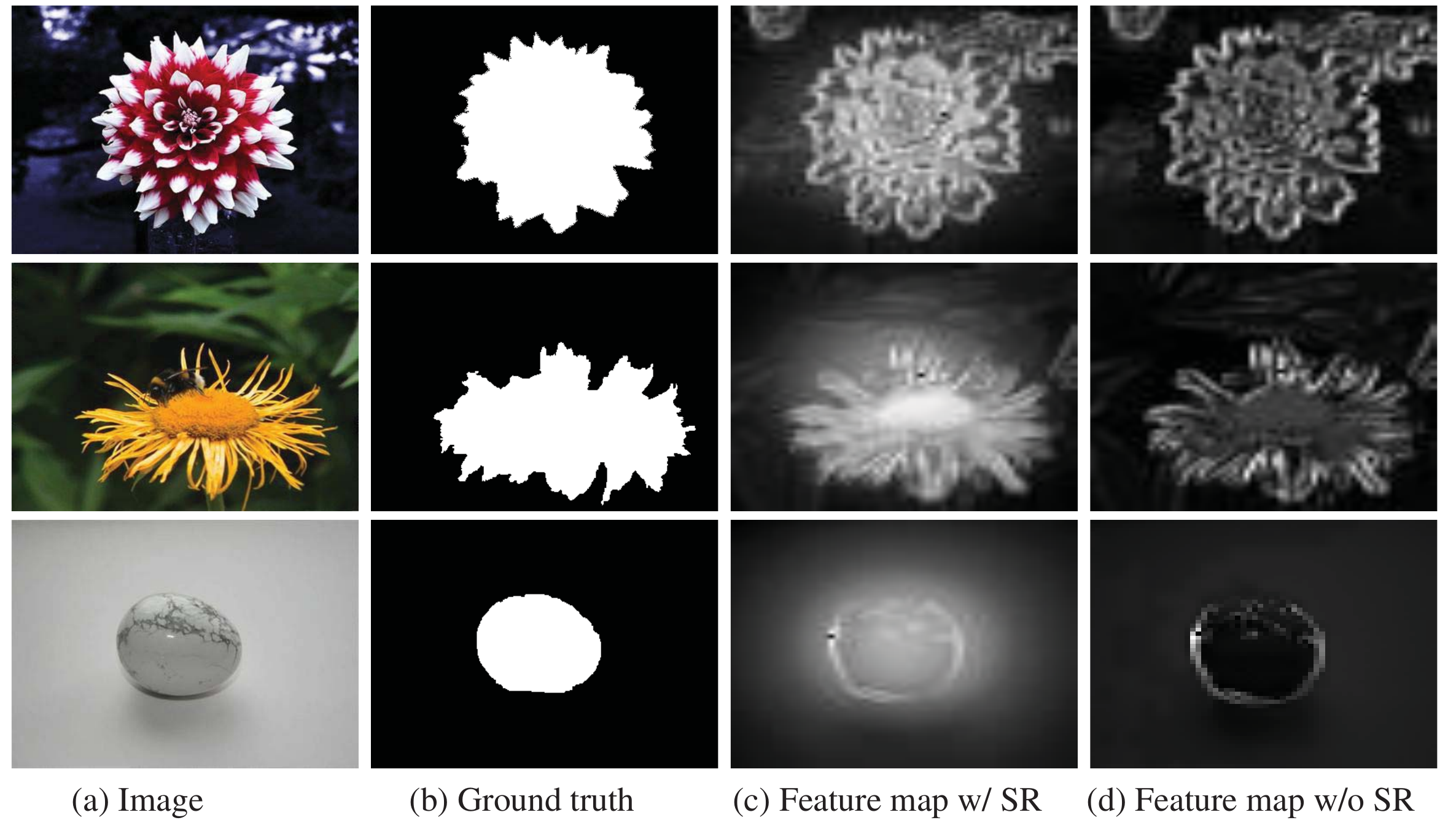}
\vspace{-6pt}
\caption{Visual comparison of feature map of the first convolutional layer with and without spatial regularization (SR). The samples are taken from the MSRA10K datasets.}
\label{fig:conv1_map}
\end{figure}

Figure \ref{fig:result_MSRA} and \ref{fig:result_gerneration} have shown that the recall of DISC is significantly higher than the previous best approaches while the precision keeps comparable with them. This indicates more salient regions are recalled by DISC, and one possible reason is that the boundary exclusion problem is alleviated.
Furthermore, we also evaluate the contribution of spatial regularization by comparing the performance of DISC with and without SR. We exclude the spatial regularization input channel in both CNNs while remaining the network architectures unchanged, and then re-train both of the CNNs.
To analyze the effectiveness of SR, we average the feature maps of the first convolutional layer, as depicted in Figure \ref{fig:conv1_map}. We find that the averaged maps without SR seem to focus only on the object contours, and SR helps to highlight the object more uniformly.
Meanwhile, the performance consistently outperforms that without spatial regularization according to three evaluation metrics, as shown in Figure \ref{fig:spatial_regularization} and Table \ref{table:spatial_regularization}.


\begin{table}[htp]
\centering
\begin{tabular}{c|c|c|c|c}
\hline
& Precision & Recall & F-measure & MAE \\
\hline\hline
w/ SR &  0.872 & 0.927 & 0.884 & 0.052 \\
\hline
w/o SR & 0.860 & 0.921 & 0.873 & 0.053 \\
\hline
\end{tabular}
\vspace{2pt}
\caption{The precision-recall with F-measure and MAE of our method trained with and without spatial regularization (SR) on the MSRA dataset.}
\vspace{-10pt}
\label{table:spatial_regularization}
\end{table}

\subsubsection{Hinge loss vs. cross entropy loss}
We finally present the experimental results that evaluate the benefit of hinge loss. For each CNN, we first connect a sigmoid layer with the top layer, and replace the hinge loss with the cross entropy loss, with the other layers left unchanged. Accordingly, the label of pixels belonging to salient object are set as 1 while others are set as 0. We then re-train two CNNs for comparison. The results on the MSRA dataset are shown in Figure \ref{fig:result_hinge_sigmoid} and Table \ref{table:svm_sigmoid}. The performance of our method trained with hinge loss is slightly better than that trained with cross entropy loss. In addition, we experimentally find a relative decrease by 8\% of MAE with hinge loss compared with cross entropy loss, which improves the visual quality of the generated saliency maps.


\begin{table}[htp]
\centering
\begin{tabular}{c|c|c|c|c}
\hline
Loss & Precision & Recall & F-measure & MAE \\
\hline\hline
hinge & 0.872 & 0.927 & 0.884 & 0.052 \\
\hline
cross entropy & 0.853 & 0.932 & 0.870 & 0.056 \\
\hline
\end{tabular}
\vspace{2pt}
\caption{The precision-recall with F-measure and MAE of our DISC model trained with hinge loss and cross entropy loss.}
\vspace{-30pt}
\label{table:svm_sigmoid}
\end{table}

\subsection{Limitation}
\vspace{-6pt}
\begin{figure}[htp]
\centering
\includegraphics[width=0.9\linewidth]{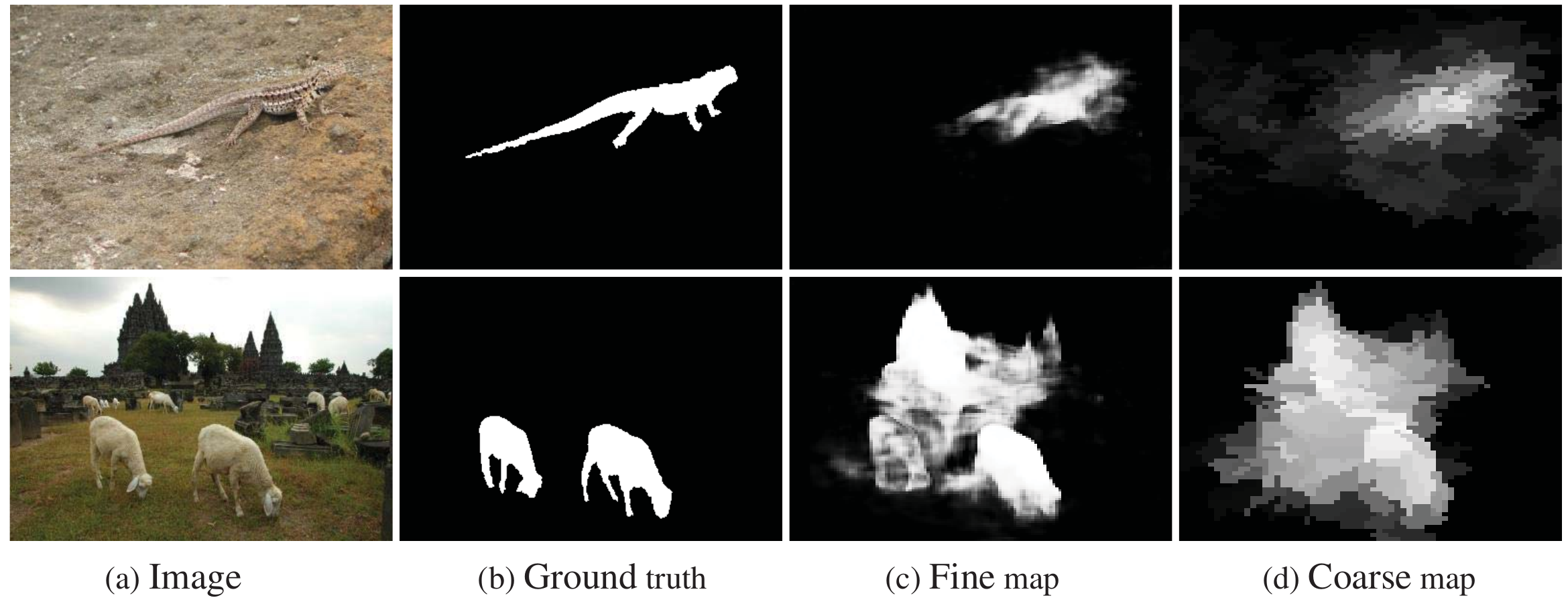}
\vspace{-6pt}
\caption{Some samples that challenge our proposed DISC framework.}
\vspace{-10pt}
\label{fig:Example_Chanllenge}
\end{figure}

In Figure \ref{fig:Example_Chanllenge}, we present some unsatisfying results generated by DISC. As discussed in \ref{subsec:SLCI}, the accuracy of fine saliency maps are influenced deeply by the quality of coarse saliency maps. In the experiment, we find that two situations will result in a poor quality coarse map. First, the coarse map fails to distinguish some foreground regions from the background when they are similar in appearance (see the first example). Second, it cannot effectively extract the foreground from cluttered background (see the second example). The SLCI will help preserve the structure of the target object, but it still cannot work well for complex background. The main limitation of our method seems to be the dependency of coarse maps. This problem could be tackled by incorporating high-level knowledge, such as object semantic shapes, to further refine the coarse saliency map.

\vspace{-6pt}
\section{Conclusion}
In this paper, we have presented an effective learning framework for accurate image saliency computing. Compared with existing image saliency models, our framework achieves superior performance without relying on any feature engineering or heuristic assumption about image saliency.
Two deep convolutional neural networks are utilized in a progressive manner to directly map the image data to detail-preserved image saliency maps.
The proposed deep architecture is very general and can be borrowed into other similar vision tasks. Extensive experimental evaluation on five public benchmarks has validated the advantages of our approach.

There are several possible directions in which we intend to extend this work. The first is to study our model in the context of generic objectness detection, which aims to fast generate a batch of hypotheses of object localizations. Second, our approach can be revised to adapt to video data, and combined with the current research of video tracking algorithms.


%

\ifCLASSOPTIONcaptionsoff
  \newpage
\fi



%

%



{
\bibliographystyle{IEEEtran}
\bibliography{reference}

\begin{thebibliography}{10}
\providecommand{\url}[1]{#1}
\csname url@samestyle\endcsname
\providecommand{\newblock}{\relax}
\providecommand{\bibinfo}[2]{#2}
\providecommand{\BIBentrySTDinterwordspacing}{\spaceskip=0pt\relax}
\providecommand{\BIBentryALTinterwordstretchfactor}{4}
\providecommand{\BIBentryALTinterwordspacing}{\spaceskip=\fontdimen2\font plus
\BIBentryALTinterwordstretchfactor\fontdimen3\font minus
  \fontdimen4\font\relax}
\providecommand{\BIBforeignlanguage}[2]{{%
\expandafter\ifx\csname l@#1\endcsname\relax
\typeout{** WARNING: IEEEtran.bst: No hyphenation pattern has been}%
\typeout{** loaded for the language `#1'. Using the pattern for}%
\typeout{** the default language instead.}%
\else
\language=\csname l@#1\endcsname
\fi
#2}}
\providecommand{\BIBdecl}{\relax}
\BIBdecl

\bibitem{itti2001computational}
L.~Itti and C.~Koch, ``Computational modelling of visual attention,''
  \emph{Nature reviews neuroscience}, vol.~2, no.~3, pp. 194--203, 2001.

\bibitem{borji2012salient}
A.~Borji, D.~N. Sihite, and L.~Itti, ``Salient object detection: A benchmark,''
  in \emph{Proc. Eur. Conf. Comput. Vis.}, Florence, Italy, Sep. 2012, pp.
  414--429.

\bibitem{li2013saliency}
X.~Li, H.~Lu, L.~Zhang, X.~Ruan, and M.-H. Yang, ``Saliency detection via dense
  and sparse reconstruction,'' in \emph{Proc. IEEE Int. Conf. Comput. Vis.},
  Sydney, Australia, Dec. 2013, pp. 2976--2983.

\bibitem{yang2013saliency}
C.~Yang, L.~Zhang, H.~Lu, X.~Ruan, and M.-H. Yang, ``Saliency detection via
  graph-based manifold ranking,'' in \emph{Proc. IEEE Conf. Comput. Vis.
  Pattern Recognit.}, Portland, OR, USA, Jun. 2013, pp. 3166--3173.

\bibitem{wang2013saliency}
Q.~Wang, Y.~Yuan, P.~Yan, and X.~Li, ``Saliency detection by multiple-instance
  learning,'' \emph{IEEE Trans. on Cybernetics}, vol.~43, pp. 660--672, Apr.
  2013.

\bibitem{wang2015pisa}
K.~Wang, L.~Lin, J.~Lu, C.~Li, and K.~Shi, ``Pisa: Pixelwise image saliency by
  aggregating complementary appearance contrast measures with edge-preserving
  coherence,'' \emph{IEEE Trans. Image Process.}, vol.~24, pp. 1057--7149, Oct.
  2015.

\bibitem{christopoulos2000jpeg2000}
C.~Christopoulos, A.~Skodras, and T.~Ebrahimi, ``The jpeg2000 still image
  coding system: an overview,'' \emph{IEEE Trans. Consumer Electronics},
  vol.~46, pp. 1103--1127, Nov. 2000.

\bibitem{han2006unsupervised}
J.~Han, K.~N. Ngan, M.~Li, and H.-J. Zhang, ``Unsupervised extraction of visual
  attention objects in color images,'' \emph{IEEE Trans. Circuits Syst. Video
  Technol.}, vol.~16, pp. 141--145, Jan. 2006.

\bibitem{lin2015discriminatively}
L.~Lin, X.~Wang, W.~Yang, and J.-H. Lai, ``Discriminatively trained and-or
  graph models for object shape detection,'' \emph{IEEE Trans. Pattern Anal.
  Mach. Intell.}, vol.~37, no.~5, pp. 959--972, 2015.

\bibitem{itti1998model}
L.~Itti, C.~Koch, and E.~Niebur, ``A model of saliency-based visual attention
  for rapid scene analysis,'' \emph{IEEE Trans. Pattern Anal. Mach. Intell.},
  vol.~20, pp. 1254--1259, Nov. 1998.

\bibitem{liu2011learning}
T.~Liu, Z.~Yuan, J.~Sun, J.~Wang, N.~Zheng, X.~Tang, and H.-Y. Shum, ``Learning
  to detect a salient object,'' \emph{IEEE Trans. Pattern Anal. Mach. Intell.},
  vol.~33, pp. 353--367, Feb. 2011.

\bibitem{zhai2006visual}
Y.~Zhai and M.~Shah, ``Visual attention detection in video sequences using
  spatiotemporal cues,'' in \emph{Proc. ACM Multimedia}, Santa Barbara, CA,
  USA, Oct. 2006, pp. 815--824.

\bibitem{achanta2009frequency}
R.~Achanta, S.~Hemami, F.~Estrada, and S.~Susstrunk, ``Frequency-tuned salient
  region detection,'' in \emph{Proc. IEEE Conf. Comput. Vis. Pattern
  Recognit.}, Miami, FL, USA, Jun. 2009, pp. 1597--1604.

\bibitem{perazzi2012saliency}
F.~Perazzi, P.~Krahenbuhl, Y.~Pritch, and A.~Hornung, ``Saliency filters:
  Contrast based filtering for salient region detection,'' in \emph{Proc. IEEE
  Conf. Comput. Vis. Pattern Recognit.}, Providence, RI, USA, Jun. 2012, pp.
  733--740.

\bibitem{shen2012unified}
X.~Shen and Y.~Wu, ``A unified approach to salient object detection via low
  rank matrix recovery,'' in \emph{Proc. IEEE Conf. Comput. Vis. Pattern
  Recognit.}, Providence, RI, USA, Jun. 2012, pp. 853--860.

\bibitem{jiang2013salient}
H.~Jiang, J.~Wang, Z.~Yuan, Y.~Wu, N.~Zheng, and S.~Li, ``Salient object
  detection: A discriminative regional feature integration approach,'' in
  \emph{Proc. IEEE Conf. Comput. Vis. Pattern Recognit.}, Portland, OR, USA,
  Jun. 2013, pp. 2083--2090.

\bibitem{krizhevsky2012imagenet}
A.~Krizhevsky, I.~Sutskever, and G.~E. Hinton, ``{ImageNet} classification with
  deep convolutional neural networks,'' in \emph{Proc. Adv. Neural Inf.
  Process. Syst.}, Harrahs and Harveys, Lake Tahoe, USA, Dec. 2012, pp.
  1097--1105.

\bibitem{huang2006large}
F.~Huang and Y.~LeCun, ``Large-scale learning with svm and convolutional
  network for generic object recognition,'' in \emph{Proc. IEEE Conf. Comput.
  Vis. Pattern Recognit.}, New York, NY, USA, Jun. 2006, pp. 284--291.

\bibitem{cheng2011global}
M.-M. Cheng, G.-X. Zhang, N.~J. Mitra, X.~Huang, and S.-M. Hu, ``Global
  contrast based salient region detection,'' in \emph{Proc. IEEE Conf. Comput.
  Vis. Pattern Recognit.}, Colorado Springs, CO, USA, Jun. 2011, pp. 409--416.

\bibitem{kadir2001saliency}
T.~Kadir and M.~Brady, ``Saliency, scale and image description,'' \emph{Int. J.
  Comput. Vis.}, vol.~45, no.~2, pp. 83--105, 2001.

\bibitem{bruce2005saliency}
N.~Bruce and J.~Tsotsos, ``Saliency based on information maximization,'' in
  \emph{Proc. Adv. Neural Inf. Process. Syst.}, Vancouver, British Columbia,
  Canada, Dec. 2005, pp. 155--162.

\bibitem{lin2014computational}
R.-J. Lin and W.-S. Lin, ``A computational visual saliency model based on
  statistics and machine learning,'' \emph{J. Vis}, vol.~14, no.~9, p.~1, 2014.

\bibitem{mai2013saliency}
L.~Mai, Y.~Niu, and F.~Liu, ``Saliency aggregation: A data-driven approach,''
  in \emph{Proc. IEEE Conf. Comput. Vis. Pattern Recognit.}, Portland, OR, USA,
  Jun. 2013, pp. 1131--1138.

\bibitem{lu2014learning}
S.~Lu, V.~Mahadevan, and N.~Vasconcelos, ``Learning optimal seeds for
  diffusion-based salient object detection,'' in \emph{Proc. IEEE Conf. Comput.
  Vis. Pattern Recognit.}, Columbus, OH, USA, Jun 2014, pp. 2790--2797.

\bibitem{liang2015towards}
X.~Liang, S.~Liu, Y.~Wei, L.~Liu, L.~Lin, and S.~Yan, ``Towards computational
  baby learning: A weakly-supervised approach for object detection,'' in
  \emph{Proc. IEEE Int. Conf. Comput. Vis.}, Santiago, Chile, Dec. 2015.

\bibitem{ding2015deep}
S.~Ding, L.~Lin, G.~Wang, and H.~Chao, ``Deep feature learning with relative
  distance comparison for person re-identification,'' \emph{Pattern Recognit.},
  vol.~48, no.~10, pp. 2993--3003, 2015.

\bibitem{deephashing}
R.~Zhang, L.~Lin, R.~Zhang, W.~Zuo, and L.~Zhang, ``Bit-scalable deep hashing
  with regularized similarity learning for image retrieval and person
  re-identification,'' \emph{IEEE Trans. Image Process.}, vol.~24, no.~12, pp.
  4766--4779, 2015.

\bibitem{liang2015human}
X.~Liang, C.~Xu, X.~Shen, J.~Yang, S.~Liu, J.~Tang, L.~Lin, and S.~Yan, ``Human
  parsing with contextualized convolutional neural network,'' in \emph{Proc.
  IEEE Int. Conf. Comput. Vis.}, Santiago, Chile, Dec. 2015.

\bibitem{lin2015deepmodel}
L.~Lin, K.~Wang, W.~Zuo, M.~Wang, J.~Luo, and L.~Zhang, ``A deep structured
  model with radius-margin bound for 3d human activity recognitin,''
  \emph{International Journal of Computer Vision}, DOI:
  10.1007/s11263-015-0876-z, 2016.

\bibitem{farabet2013learning}
C.~Farabet, C.~Couprie, L.~Najman, and Y.~LeCun, ``Learning hierarchical
  features for scene labeling,'' \emph{IEEE Trans. Pattern Anal. Mach.
  Intell.}, vol.~35, pp. 1915--1929, Aug. 2013.

\bibitem{sermanet2013pedestrian}
P.~Sermanet, K.~Kavukcuoglu, S.~Chintala, and Y.~LeCun, ``Pedestrian detection
  with unsupervised multi-stage feature learning,'' in \emph{Proc. IEEE Conf.
  Comput. Vis. Pattern Recognit.}, Portland, OR, USA, Jun. 2013, pp.
  3626--3633.

\bibitem{wang2014deep}
X.~Wang, L.~Zhang, L.~Lin, Z.~Liang, and W.~Zuo, ``Deep joint task learning for
  generic object extraction,'' in \emph{Proc. Adv. Neural Inf. Process. Syst.},
  Montreal, Quebec, Canada, Dec. 2014, pp. 523--531.

\bibitem{sun2013deep}
Y.~Sun, X.~Wang, and X.~Tang, ``Deep convolutional network cascade for facial
  point detection,'' in \emph{Proc. IEEE Conf. Comput. Vis. Pattern Recognit.},
  Portland, OR, USA, Jun. 2013, pp. 3476--3483.

\bibitem{eigen2014depth}
D.~Eigen, C.~Puhrsch, and R.~Fergus, ``Depth map prediction from a single image
  using a multi-scale deep network,'' in \emph{Proc. Adv. Neural Inf. Process.
  Syst.}, Montreal, Quebec, Canada, Dec. 2014, pp. 2366--2374.

\bibitem{he2015supercnn}
S.~He, R.~W. Lau, W.~Liu, Z.~Huang, and Q.~Yang, ``Supercnn: A superpixelwise
  convolutional neural network for salient object detection,'' \emph{Int. J.
  Comput. Vis.}, vol. 115, no.~3, pp. 1--15, 2015.

\bibitem{li2015visual}
G.~Li and Y.~Yu, ``Visual saliency based on multiscale deep features,''
  \emph{arXiv preprint arXiv:1503.08663}, 2015.

\bibitem{wang2015deep}
L.~Wang, H.~Lu, X.~Ruan, and M.-H. Yang, ``Deep networks for saliency detection
  via local estimation and global search,'' in \emph{Proc. IEEE Conf. Comput.
  Vis. Pattern Recognit.}, Boston, MA, USA, Jun. 2015, pp. 3183--3192.

\bibitem{russakovsky2014imagenet}
O.~Russakovsky, J.~Deng, H.~Su, J.~Krause, S.~Satheesh, S.~Ma, Z.~Huang,
  A.~Karpathy, A.~Khosla, M.~Bernstein \emph{et~al.}, ``Imagenet large scale
  visual recognition challenge,'' \emph{Int. J. Comput. Vis.}, vol. 115, no.~3,
  pp. 211--252, 2014.

\bibitem{liu2015predicting}
N.~Liu, J.~Han, D.~Zhang, S.~Wen, and T.~Liu, ``Predicting eye fixations using
  convolutional neural networks,'' in \emph{Proc. IEEE Conf. Comput. Vis.
  Pattern Recognit.}, Boston, MA, USA, Jun. 2015, pp. 362--370.

\bibitem{vig2014large}
E.~Vig, M.~Dorr, and D.~Cox, ``Large-scale optimization of hierarchical
  features for saliency prediction in natural images,'' in \emph{Proc. IEEE
  Conf. Comput. Vis. Pattern Recognit.}, Columbus, OH, USA, Jun. 2014, pp.
  2798--2805.

\bibitem{kummerer2014deep}
M.~K{\"u}mmerer, L.~Theis, and M.~Bethge, ``Deep {G}aze {I}: Boosting saliency
  prediction with feature maps trained on imagenet,'' \emph{arXiv preprint
  arXiv:1411.1045}, 2014.

\bibitem{shen2014learning}
C.~Shen and Q.~Zhao, ``Learning to predict eye fixations for semantic contents
  using multi-layer sparse network,'' \emph{Neurocomputing}, vol. 138, pp.
  61--68, Aug. 2014.

\bibitem{liu2011entropy}
M.-Y. Liu, O.~Tuzel, S.~Ramalingam, and R.~Chellappa, ``Entropy rate superpixel
  segmentation,'' in \emph{Proc. IEEE Conf. Comput. Vis. Pattern Recognit.},
  Colorado Springs, CO, USA, Jun. 2011, pp. 2097--2104.

\bibitem{long2015fully}
J.~Long, E.~Shelhamer, and T.~Darrell, ``Fully convolutional networks for
  semantic segmentation,'' in \emph{Proc. IEEE Conf. Comput. Vis. Pattern
  Recognit.}, Boston, MA, USA, Jun. 2015, pp. 3431--3440.

\bibitem{xu2013soft}
X.~Xu, I.~W. Tsang, and D.~Xu, ``Soft margin multiple kernel learning,''
  \emph{IEEE Trans. Neural Netw. Learn. Syst.}, vol.~24, pp. 749--761, May
  2013.

\bibitem{alpert2007image}
S.~Alpert, M.~Galun, R.~Basri, and A.~Brandt, ``Image segmentation by
  probabilistic bottom-up aggregation and cue integration,'' in \emph{Proc.
  IEEE Conf. Comput. Vis. Pattern Recognit.}, Minneapolis, MN, USA, 2007, pp.
  1--8.

\bibitem{yan2013hierarchical}
Q.~Yan, L.~Xu, J.~Shi, and J.~Jia, ``Hierarchical saliency detection,'' in
  \emph{Proc. IEEE Conf. Comput. Vis. Pattern Recognit.}, Portland, OR, USA,
  Jun. 2013, pp. 1155--1162.

\bibitem{zou2013segmentation}
W.~Zou, K.~Kpalma, Z.~Liu, and J.~Ronsin, ``Segmentation driven low-rank matrix
  recovery for saliency detection,'' in \emph{Proc. Brit. Mach. Vis. Conf.},
  Bristol, Britain, Sep. 2013, pp. 1--13.

\bibitem{cheng2014salientshape}
M.-M. Cheng, N.~J. Mitra, X.~Huang, and S.-M. Hu, ``Salientshape: Group
  saliency in image collections,'' \emph{The Visual Computer}, vol.~30, no.~4,
  pp. 443--453, 2014.

\bibitem{jia2014caffe}
Y.~Jia, E.~Shelhamer, J.~Donahue, S.~Karayev, J.~Long, R.~Girshick,
  S.~Guadarrama, and T.~Darrell, ``Caffe: Convolutional architecture for fast
  feature embedding,'' in \emph{Proc. ACM Multimedia}, Orlando, FL, USA, Nov.
  2014, pp. 675--678.

\bibitem{jiang2011automatic}
H.~Jiang, J.~Wang, Z.~Yuan, T.~Liu, N.~Zheng, and S.~Li, ``Automatic salient
  object segmentation based on context and shape prior.'' in \emph{Proc. Brit.
  Mach. Vis. Conf.}, Dundee, Britain, Aug./Sep. 2011, pp. 1--12.

\bibitem{jiang2013saliency}
B.~Jiang, L.~Zhang, H.~Lu, C.~Yang, and M.-H. Yang, ``Saliency detection via
  absorbing markov chain,'' in \emph{Proc. IEEE Int. Conf. Comput. Vis.},
  Sydney, Australia, Dec. 2013, pp. 1665--1672.

\bibitem{harel2006graph}
J.~Harel, C.~Koch, and P.~Perona, ``Graph-based visual saliency,'' in
  \emph{Proc. Adv. Neural Inf. Process. Syst.}, Vancouver, British Columbia,
  Canada, Dec. 2006, pp. 545--552.

\end{thebibliography}
}




\vspace{-55pt}
\begin{IEEEbiography}[{\includegraphics[width=1in,height=1.25in,clip,keepaspectratio]{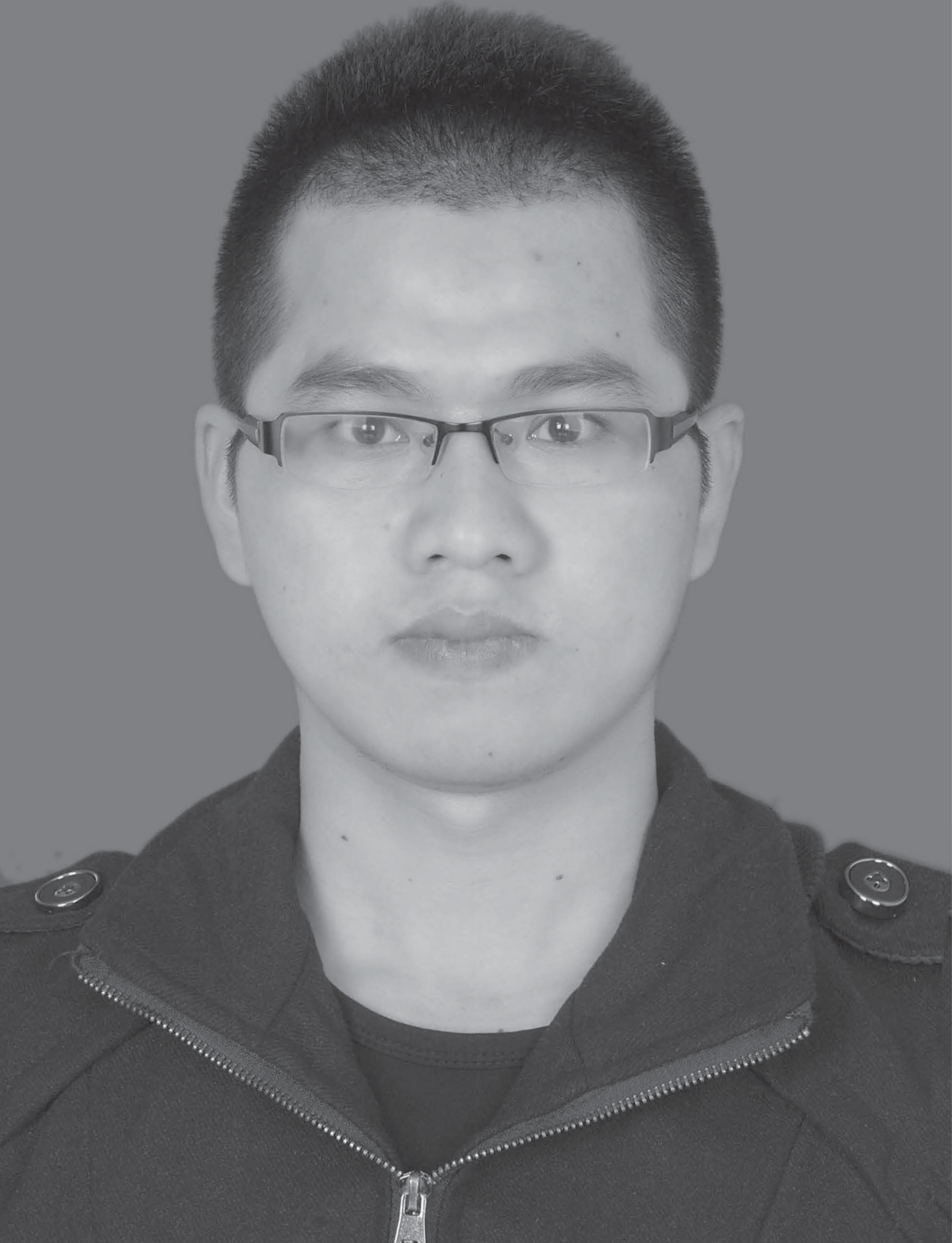}}]{Tianshui Chen} received the B.E. degree from School of Information and Science Technology, Sun Yat-sen University, Guangzhou, China, in 2013, where he is currently pursuing the Ph.D. degree in computer science with the School of Data and Computer Science. His current research interests include computer vision and machine learning.
\end{IEEEbiography}

\vspace{-45pt}
\begin{IEEEbiography}[{\includegraphics[width=1in,height=1.25in,clip,keepaspectratio]{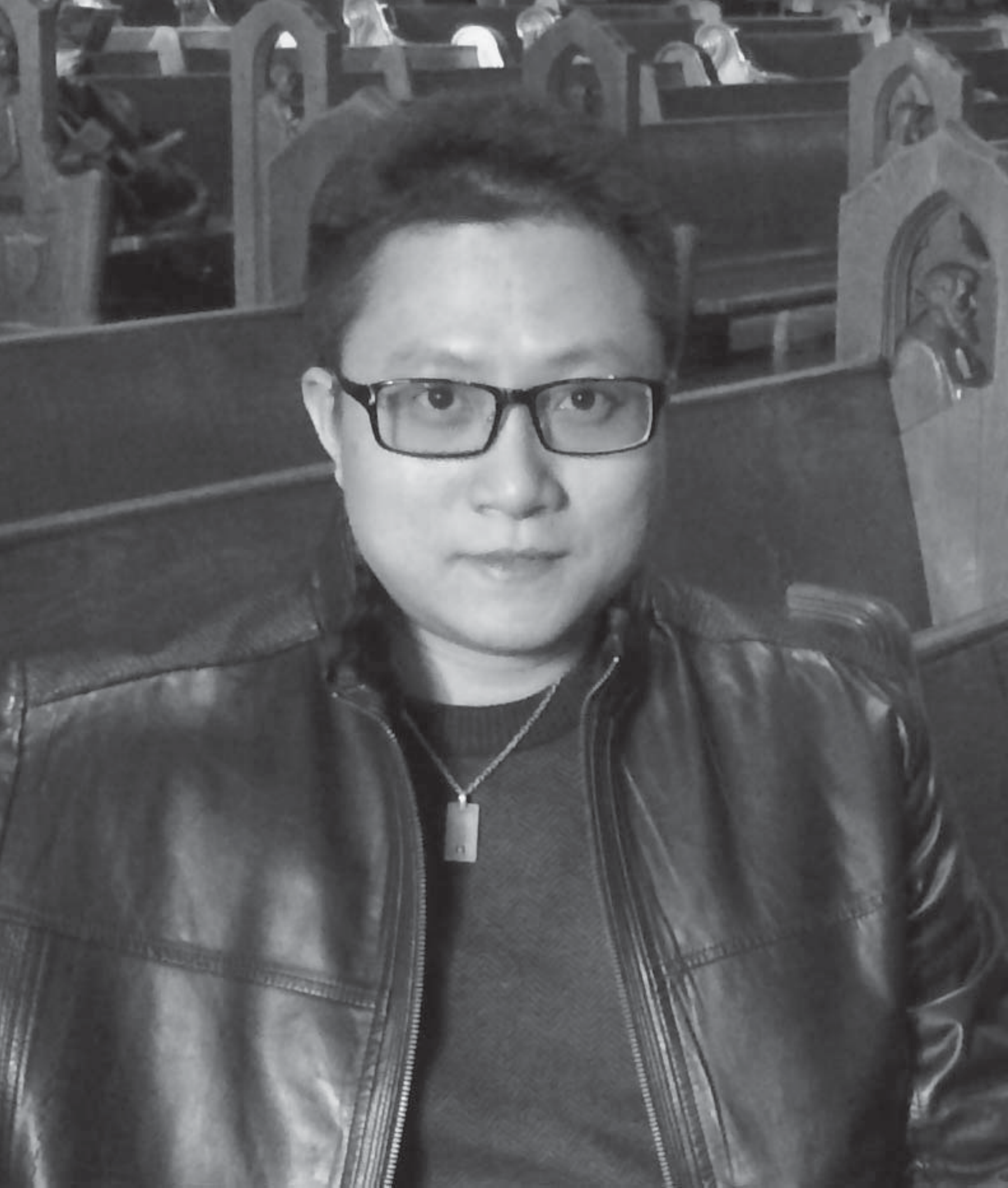}}]{Liang Lin} is a Professor with the School of Data and Computer Science, Sun Yat-sen University (SYSU), China. He received the B.S. and Ph.D. degrees from the Beijing Institute of Technology (BIT), Beijing, China, in 1999 and 2008, respectively. From 2006 to 2007, he was a joint Ph.D. student with the Department of Statistics, University of California, Los Angeles (UCLA). His Ph.D. dissertation was nominated by the China National Excellent PhD Thesis Award in 2010. He was a Post-Doctoral Research Fellow with the Center for Vision, Cognition, Learning, and Art of UCLA.
Prof. Lin's research focuses on new models, algorithms and systems for intelligent processing and understanding of visual data. He has authorized or co-authorized more than 80 papers in top tier academic journals and conferences. He has served as an associate editor for journal Neurocomputing and The Visual Computer, and a guest editor for Pattern Recognition. He was supported by several promotive programs or funds for his works, such as Guangdong NSFs for Distinguished Young Scholars in 2013. He received the Best Paper Runners-Up Award in ACM NPAR 2010, Google Faculty Award in 2012, and Best Student Paper Award in IEEE ICME 2014.
\end{IEEEbiography}

\vspace{-45pt}
\begin{IEEEbiography}[{\includegraphics[width=1in,height=1.25in,clip,keepaspectratio]{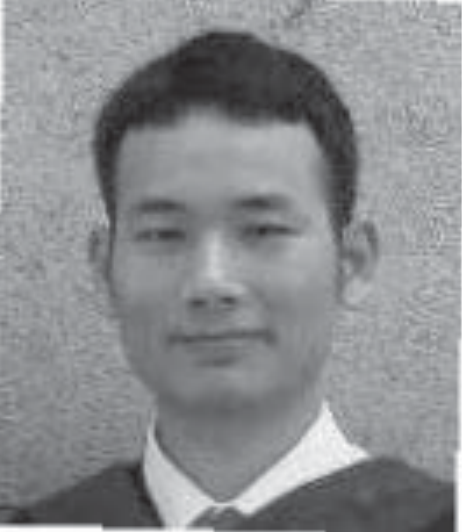}}]{Lingbo Liu} received the B.E. degree from the School of Software, Sun Yat-sen University, Guangzhou, China, in 2015. He is currently pursuing the M.Sc. degree in computer science with the School of Data and Computer Science. His research interest includes computer vision, machine learning, and parallel computation.
\end{IEEEbiography}

\vspace{-450pt}
\begin{IEEEbiography}[{\includegraphics[width=1in,height=1.25in,clip,keepaspectratio]{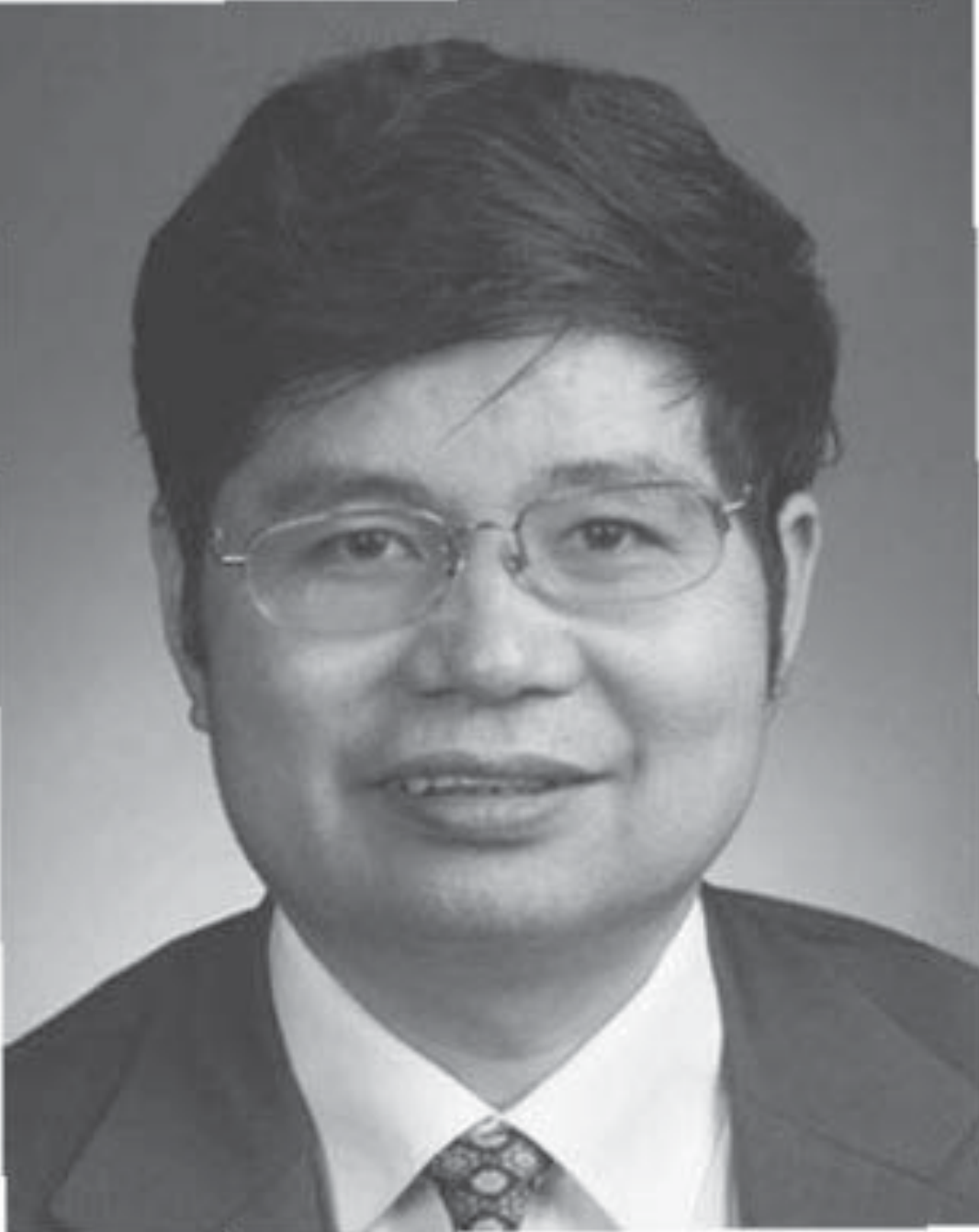}}]{Xiaonan Luo} is a professor of the School of Information and Science Technology at Sun Yat-sen University. He also serves as Director of the National Engineering Research Center of Digital Life. His research interests include Computer Vision, Image Processing, and Computer Graphics \& CAD.
\end{IEEEbiography}

\vspace{-450pt}
\begin{IEEEbiographynophoto}{Xuelong Li}
(M'02-SM'07-F'12) is a full professor with the Center for OPTical IMagery Analysis and Learning (OPTIMAL), State Key Laboratory of Transient Optics and Photonics, Xi'an Institute of Optics and Precision Mechanics, Chinese Academy of Sciences, Xi'an 710119, Shaanxi, P. R. China.
\end{IEEEbiographynophoto}

\end{document}